\documentclass[journal]{IEEEtran}
\pdfoutput=1
\usepackage{cite}
\usepackage{multirow}

\ifCLASSINFOpdf
\else
\fi
\usepackage{amssymb}
\usepackage{amsmath}
\DeclareMathOperator*{\argmax}{argmax}

\usepackage{algorithmic}
\usepackage{algorithm}

\usepackage{array}
\usepackage{slashbox}
\usepackage{threeparttable}

\usepackage{graphics}
\usepackage{graphicx}
\ifCLASSOPTIONcompsoc
  \usepackage[caption=false,font=normalsize,labelfont=sf,textfont=sf]{subfig}
\else
  \usepackage[caption=false,font=footnotesize]{subfig}
\fi
\usepackage{color}
\begin{document}
%
\title{Joint Hierarchical Category Structure Learning and Large-Scale Image Classification}
%
%
%

\author{Yanyun Qu,~\IEEEmembership{Member,~IEEE,}
        Li Lin, Fumin Shen, Chang Lu,\\
        Yang Wu,
        Yuan Xie,~\IEEEmembership{Member,~IEEE,}
        Dacheng Tao,~\IEEEmembership{Fellow,~IEEE,}
\thanks{\footnotesize Yanyun Qu, Li Lin, and Chang Lu are with the Department
of Computer Science, Xiamen University, Xiamen, 361005, China
(e-mail: yyqu@xmu.edu.cn). }
\thanks{\footnotesize Fumin Shen is with the School of Computer Science and Engineering, University of Electric Science and technology of China, Chengdu, 611731, China (fumin.shen@gmail.com).}
\thanks{\footnotesize Yang Wu is with Institute for Research Initiatives, Nara Institute of Science and Technology, Japan  (yangwu@rsc.naist.jp).}
\thanks{\footnotesize Yuan Xie is with Research Center of Precision Sensing and Control
Institute of Automation, Chinese Academy of Sciences, Beijing, 100190, China (email:yuan.xie@ia.ac.cn).}
\thanks{\footnotesize Dacheng Tao is with the Centre for Artificial Intelligence and the Faculty of Engineering and Information Technology, University of Technology Sydney, 81 Broadway Street, Ultimo, NSW 2007, Australia (email: dacheng.tao@uts.edu.au)
}
}

%
%

\markboth{IEEE Transaction on Image Processing}%
{Shell \MakeLowercase{\textit{et al.}}: Bare Demo of IEEEtran.cls for IEEE Journals}
%



\maketitle

\begin{abstract}
We investigate the scalable image classification problem with a large number of categories. Hierarchical visual data structures are helpful for improving the efficiency and performance of large-scale multi-class classification. We propose a novel image classification method based on learning hierarchical inter-class structures. Specifically, we first design a fast algorithm to compute the similarity metric between categories, based on which a visual tree is constructed by hierarchical spectral clustering. Using the learned visual tree, a test sample label is efficiently predicted by searching for the best path over the entire tree. The proposed method is extensively evaluated on the ILSVRC2010 and Caltech 256 benchmark datasets. Experimental results show that our method obtains significantly better category hierarchies than other state-of-the-art visual tree-based methods and, therefore, much more accurate classification.
\end{abstract}


%
\IEEEpeerreviewmaketitle

\section{Introduction}\label{sec:introduction}
%
%
%
%
\IEEEPARstart{G}{reat} progresses have been witnessed in image classification \cite{00item, 01item, 02item, 03item, 04item, 05item, 06item, 07item, 51item, 52item} in recent years. Especially, large-scale image classification has achieved remarkable developments \cite{53item, 54item, 55item, 56item, 58item}. Nevertheless, most state-of-the-art methods have the following two limitations: 1) inter-class taxonomic relationships are neglected; thus, the hierarchical structure of multiple classes cannot be generated and visualized; and 2) multi-class classification decision-making is ``flat'', making computation inefficient.

With respect to visual hierarchical relationships, as the explosive rise in diverse social media data extends beyond direct administration by individual users, a proper hierarchical structure could make it easier for users to capture the distribution of image data such that they can effectively manage and organize their data. Moreover, it is natural to organize data according to their relationships and form a hierarchical structure. For example, ImageNet  \cite{1item} is organized hierarchically according to a high-level semantic lexical database called WordNet \cite{2item}, which classifies objects in the natural world according to phylum, class, order, family, genus, and species, i.e., a well-established hierarchy.

With respect to multi-class classification, most methods simply directly adopt a flat scheme, i.e., one-vs.-all or one-vs.-one classifiers, making prediction time-consuming. For N classes, one-vs.-all needs to compute $N$ classifiers and one-vs.-one needs to compute $N(N-1)/2$ classifiers when predicting a query image. When $N$ is large, the two flat methods are not efficient. Thus, we seek hierarchical structure to improve the efficiency of multi-class prediction which requires $O(log_{K}N)$ classifiers for a tree with $K$ branches in each layer. Moreover, real-world object classes tend to have strong hierarchical relationships, and it is usually easier for humans to distinguish coarse-level categories than fine-grained subcategories. Given a hierarchical structure, classification can be performed in a coarse-to-fine manner, which improves prediction efficiency and accuracy. To this end, we propose an efficient multi-class classification framework based on hierarchical category structure learning.

The first challenge is how to apply a hierarchical structure to visual data. Hierarchical classification methods typically depend on a given hierarchical structure, e.g., WordNet for ImageNet. However, the construction of ImageNet was demanding and time-consuming. In real world scenarios, knowledge about how to organize data hierarchically is often limited; we usually only know some coarse and obscure dataset cues. Thus, it is difficult to utilize high-level semantics to construct a hierarchical structure. Moreover, there is no evidence to suggest that class prediction according to hierarchical semantics improves performance; the classification accuracy can be low even when a hierarchical semantic structure is used \cite{3item,4item}. This is probably due to the semantic gap between high-level semantics and low-level features. Thus, we propose making hierarchical classification dependent on a visual tree constructed using visual features but not predefined rules.

Performing hierarchical inference based on the visual hierarchical structure is also challenging. Greedy learning is a typical way to solve classification prediction using visual tree models \cite{3item,5item,6item}. However, while relatively intuitive, the greedy approach does not prevent error propagation; that is, if a mistake is made in an intermediate node, the prediction result is then destined to be wrong. To overcome this drawback, we transform the problem of class prediction into a task of finding the optimal path of a visual tree by maximizing a joint probability.

The main contributions of our approach are as follows:
\begin{enumerate}
  \item We propose a fast approach for computing the between-category similarity metric, based on which hierarchical spectral clustering is used to construct the visual tree.
  \item To avoid error propagation, we transform the class prediction problem into a path-searching problem using a novel method that we call \emph{N-best path}. \emph{N-best path} is an approximation of the optimal path solved by the maximum joint probability of candidate paths. Rather than finding only one path, we retain candidate paths corresponding to the top $N$ largest joint probabilities. Compared to traditional greedy learning methods, the \emph{N-best path} algorithm effectively avoids error propagation and improves prediction efficiency.
  \item The proposed classification framework is extensively evaluated with respect to different image representations including hand-crafted features and recently developed deep features. Furthermore, we compare the proposed approach with several other visual tree-based algorithms on two large datasets: ILSVRC2010 and Caltech 256. Our method produces significantly better category hierarchies and thus improves classification accuracy compared to the previous state-of-the-art methods.
\end{enumerate}

The remainder of this paper is organized as follows. We introduce related work in Section \ref{sec:related work}. In Section \ref{sec:algorithm overview}, we construct a visual tree model and detail the \emph{N-best path} algorithm for label inference depending on the visual tree. Experimental results are presented in Section \ref{sec:experimental results}, and we conclude in Section \ref{sec:conclusions}.

\section{Related Work}\label{sec:related work}

\subsection{Hierarchical learning}
There are two groups of hierarchical learning approaches: taxonomy-related methods and taxonomy-independent methods. Motivated by the success of taxonomies in web organization and text document categorization, many computer vision researchers have utilized taxonomies to organize large-scale image collections or improve visual system performance. For instance, Li et al. \cite{1item} constructed ImageNet according to WordNet, a semantic hierarchy taxonomy unrelated to visual effects. Although WordNet has been widely applied to image classification \cite{4item}, the visual attributes are always ignored.

There are precedents that learning visual hierarchical structures can be helpful for image classification \cite{11item,12item}. Sivic et al. \cite{13item} used a hierarchical Latent Dirichlet Allocation (hLDA) on Bag Of Word (BOW) \cite{8item} with SIFT local features to discover a hierarchical structure from unlabeled images, which simultaneously facilitated image classification and segmentation, while Bart et al. \cite{14item} utilized an unsupervised Bayesian model on BOW with color-space histograms to learn a tree structure. Both methods were tested on moderate-scale datasets; their performance on large-scale image data is less clear. Moreover, they focused on image classification but did not visualize the hierarchical inter-class relationship. To do so, some researchers \cite{5item,10item,15item} have built hierarchical models based on confusion matrices obtained or computed by the output of image categorization or object classification using $N$ one-vs.-all SVM classifiers. Griffin et al. \cite{15item} constructed a binary branch tree to improve visual categorization, Bengio et al. \cite{5item} built a label-embedding tree for multi-class classification, while Liu et al. \cite{10item} constructed a probabilistic label tree for large-scale classification. Gao, et al. \cite{7item} built the relaxed hiearchical structure which allows the confusion classes belong to more than one node. However, hierarchical learning methods based on confusion matrices suffer from two main limitations: 1) computation of the confusion matrix using a one-vs.-all SVM is time consuming; and 2) the confusion matrix may not be reliable due to unbalanced training data. Visual trees constructed by clustering produce an intuitive hierarchical structure \cite{3item,6item,17item,18item} and have attracted more and more attention. Zhou et al. \cite{6item} utilized AP clustering and Lei et al. \cite{3item} implemented spectral clustering to construct visual trees. Although results were promising with these methods, there is still plenty of room for improvement in hierarchical learning.

There are three important components to hierarchical learning: image representation, hierarchical structure construction, and multi-class classification inference. The greedy learning method is typically utilized for class prediction. Most hierarchical classification approaches \cite{5item,15item,16item,18item, 42item,43item} make predictions in each layer by maximizing the classification probability. However, as noted above, inferences from greedy learning do not prevent error propagation. In contract, our method provides a distribution of hierarchical memberships for image categories based on spectral clustering, in which the best path algorithm is developed to avoid error propagation. The closest related work is \cite{4item}, where the best path is learned by the structured SVM, leading to high computational complexity. We make classification predictions based on the best path algorithm depending on the hierarchical structure.

\subsection{Image representation}
Many image representation methods \cite{04item, 06item} have been used in computer vision with the BOW model, one of the most popular tools for image representation in image classification \cite{9item,19item, 20item}, image annotation \cite{05item} and image segmentation. The main advantage of the BOW model is that it is universal for image classification, meaning that it can represent generic classes of objects other than those for special object recognition (e.g., Haar-like features are especially effective for face detection). Usually, a BOW model includes three important components: 1) local feature extraction, 2) visual feature encoding, and 3) the classifier design. Local feature extraction is a prerequisite for image classification. The more discriminative the features, the better the image classification performance. Most common local features such as SIFT \cite{21item} and HOG \cite{22item} are carefully engineered.

In view of the need for a visual dictionary and encoding, K-means is traditionally used to construct a visual dictionary and the cluster centers are treated as the visual words. K-means is an unsupervised method, so the visual dictionary lacks discriminatory power. Therefore, efforts have been made to encode discriminative features. Yang et al. \cite{23item} used sparse representations to encode the features, leading to improvements in dictionary learning for multi-class classification with large numbers of classes \cite{6item,42item,43item,44item,45item}. More advanced methods have emerged over recent years such as local-constrained linear coding \cite{24item}, super-vector coding \cite{25item}, and Fisher vectors \cite{26item}. Although these methods have generally improved the discrimination of visual features, they depend on experts. More flexible and effective features are required for large-scale image classification.

Deep features learned by deep learning have become more prevalent over recent years and can be obtained in an end-to-end manner without much human intervention. They have been highly successful for audio and text recognition. Lecun et al. \cite{27item} designed a convolutional neural network (CNN) for object recognition that combined feature extraction with classifier design. Krizhevsky et al.  \cite{28item} constructed a deep CNN for large-scale image classification, while Christian improved GoogLeNet and developed the Inception V3 model \cite{46item}. However, there is a paucity of literature on how deep features influence hierarchical learning. Here, we bridge this gap by exploring the effect of deep features on hierarchical learning.

\section{Hierarchical Learning Algorithm Overview}\label{sec:algorithm overview}
\begin{figure*}[!t]
\centering
\includegraphics[width=\textwidth]{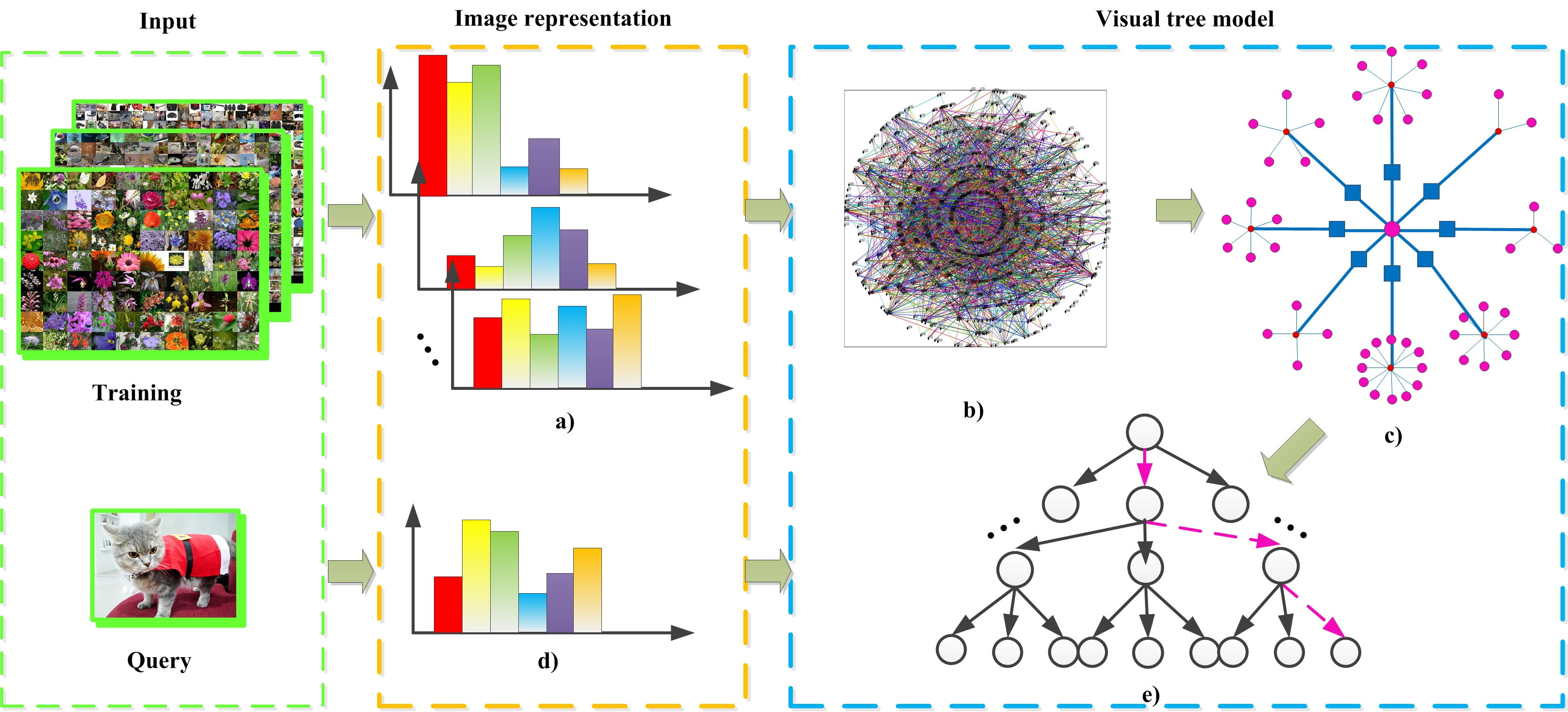}
\caption{The hierarchical learning framework. a) Image representation. b) Affinity network construction for similarity comparison between  two categories based on a similarity metric. c) Visual tree construction via hierarchical spectral clustering. d) Image representation for a query image. e) Label inference according to the visual tree model.}
\label{fig:framework1}
\end{figure*}

The framework of our approach is shown in  \figurename \ref{fig:framework1}. It has two components: visual tree construction and class prediction. In the former, an image representation is made for each image, after which an affinity matrix is computed that measures the inter-class similarity. The hierarchical category structure is then found according to the affinity matrix by spectral clustering. A visual tree model is made by assigning a weight to each edge. During testing, a query image is first represented before a prediction being made according to the visual tree model. We detail our approach below.

\subsection{Visual Tree Construction}
In this subsection, we detail how to construct the visual tree. There are two main components to our visual tree construction algorithm: 1) the similarity metric between two categories, and 2) hierarchical clustering for visual tree construction. Aligning the inter-class semantic similarity with the inter-class visual similarity is still an unsolved issue in the multimedia and computer vision communities. Human perceptual factors may be important for designing a more suitable cross-modal alignment framework. Some methods \cite{5item,10item} have constructed the visual tree according to the confusion matrix obtained by training a one-vs.-all classifier for all $N$ classes. However, these methods are computationally demanding.

\textbf{Affinity matrix computation.} In this paper, we compute an affinity matrix based on a new inter-class distance metric to construct a visual tree. This produces an algorithm that is much faster than those using traditional pairwise distances. Moreover, the proposed inter-class distance metric can be used to illustrate the relationship between two types of inter-class distance metric: the distance based on two class means \cite{6item,30item} and the distance based on the pairwise distance of two classes \cite{3item,17item}.

Suppose that there are $N$ image categories $\{C_{1},C_{2},\cdots,C_{N}\}$, and the $ith$ image category $C_{i}$ contains $N_{i}$ images represented by the features $\{I_{l}^{i}\}_{l=1}^{N_{i}}$. The similarity metric based on the pairwise distance between two classes is formulated as,
\begin{equation}\label{equ 1}
    dis(C_i,C_j) = sqrt(\frac 1{N_iN_j}\sum_s\sum_t\parallel I_s^i-I_t^j\parallel ^2)
\end{equation}

We take the norm operation as a unit, and (\ref{equ 1}) requires $N_{i}N_{j}$ norm operations. To reduce the computation of (\ref{equ 1}), we infer it as,
\begin{equation}\label{equ 2}
    dis^2(C_i,C_j) = \frac 1{N_iN_j}\sum_s\sum_t\parallel(Q_i-\Delta I_s^i)-(Q_j-\Delta I_t^j)\parallel^2
\end{equation}
where $\Delta I_{s}^{i} =Q_{i}-I_{s}^{i}$ is the difference between the image $I_{s}^{i}$ and the mean of their class $Q_{i}$, where $Q_{i}=\frac{1}{N_{i}}\sum_{l=1}^{N_{i}} I_{l}^{i}$. Furthermore, we substitute the property $\sum_{s=1}^{N_{i}}\Delta I_{s}^{i}=0$ into (\ref{equ 2}), obtaining a new distance formula,

\begin{equation}\label{equ 3}
\begin{split}
    &dis^2(C_i,C_j)  \\
    &= \frac 1{N_iN_j}\sum_{s=1}^{N_i}\sum_{t=1}^{N_j}\parallel(Q_i-Q_j)-(\Delta I_s^i-\Delta I_t^j)\parallel^2  \\
    &=\parallel Q_i-Q_j\parallel^2 - \frac 2{N_iN_j}\|Q_i-Q_j\|\sum_{s=1}^{N_i}\sum_{t=1}^{N_j}(\Delta I_s^i-\Delta I_t^j)  \\
    & + \frac 1{N_iN_j}\sum_{s=1}^{N_i}\sum_{t=1}^{N_j}(\Delta I_s^i-\Delta I_t^j)^2  \\
    &=\|Q_i-Q_j\|^2+\frac 1{N_i}\sum_{s=1}^{N_i}(\Delta I_s^i)^2+\frac 1{N_j}\sum_{t=1}^{N_j}(\Delta I_t^i)^2  \\
    &=\|Q_i-Q_j\|^2+\sigma_i^2+\sigma_j^2 \\
\end{split}
\end{equation}
where $\sigma_{i}^{2}=\frac{1}{N_{i}} \sum_{l=1}^{N_{i}}\|I_{l}^{i}-Q_{i}\|^{2}$ is the square of the variance of the category $C_{i}$. (\ref{equ 3}) requires only $N_{i}+N_{j}+1$ norm operations which are much less than those needed by (\ref{equ 1}). Considering that the variance can be pre-computed for each category, the computational cost is the same as that for the distance between class means.

Though the inference of (\ref{equ 3}) is simple, it obviously improves
 the computation of the inter-class distance. More importantly, it illustrates the core of the inter-class distance. The proposed similarity metric is related to both between-class scatter and within-class scatter. If the two centers of pairwise classes are closer and the divergence of the two classes are smaller, the similarity of two classes is bigger.

We construct a visual affinity graph by using the similarity metric between categories. Hierarchical spectral clustering \cite{47item} is then applied to construct a visual tree. The element of the affinity matrix is computed as,
\begin{equation}\label{equ 4}
    A_{ij}= exp(-\frac {dis(C_i,C_j)}{\delta_{ij}})
\end{equation}
where $\delta_{ij}$ is the self-tuning parameter according to \cite{20item}.

\textbf{Visual tree construction.} We adopt the top-down strategy to construct the visual tree. Each node in the tree is partitioned recursively during the construction procedure. The root node $v$ is set to $0$ and its depth is set to $1$. Since the root node contains all categories, spectral clustering is implemented on it based on the entire affinity matrix, and it is divided into K groups that form $K$ child nodes. The depth of the child nodes is set to $2$. Each child node contains some categories, and spectral clustering is used on a child node according to the affinity matrix corresponding to the categories contained in this node. The operation is run recursively until any of the following rules are met:
\begin{enumerate}
\item The current node is a leaf node.
\item The number of branches in the current node is less than $K$.
\item The depth is the maximum depth $L$.
\end{enumerate}

\begin{algorithm}
\caption{Construction of a visual tree via spectral clustering}
\label{algorithm 1}
\begin{algorithmic}

\REQUIRE {Training  data of $N$ image categories, $C(v)$ is the set of classes contained in the node $v$, branching factor $K$ and maximum depth $L$}
\ENSURE {Hierarchical structure of a visual tree}

//Affinity matrix construction \;
\FOR{$i=1$ \TO $N$}
    \STATE Compute the mean vector $Q_i$ and variance $\sigma_i$ of the $i$th category
\ENDFOR

\FOR{$i=1$ \TO $N$}
    \FOR{$j=1$ \TO $N$}
        \IF{$i=j$}
            \STATE $A_{i,j}\leftarrow 1$
        \ELSE
            \STATE Compute the element $A_{i,j}$ according to (\ref{equ 4})
        \ENDIF
    \ENDFOR
\ENDFOR

//Hierarchical clustering for a visual tree construction \;
\STATE Make a root node with depth $1$
\FOR{$d=1$ \TO $L$}
    \FORALL{$v$ such that depth($v$) = $d$}
        \IF{$|C(v)|< K $}
            \STATE Generate $|C(v)|$ nodes as the children node (leaf nodes) where each node contains only one category
        \ELSE
            \STATE Partition the related label set $C(v)$ into $K$ disjoint subsets by spectral clustering based on the affinity matrix $A$
            \STATE The $i$th children node which contains the class set $C_{v}^{i}$, $i=1,2,\cdots,K$,  with $\bigcup\limits_{i\in C(v)}C_{v}^{i}=C(v)$, and $C_{v}^{i}\cap C_{v}^{j}=\emptyset$, $i\neq j$
        \ENDIF
    \ENDFOR
\ENDFOR

\end{algorithmic}
\end{algorithm}

\begin{figure}[!t]
\centering
\includegraphics[width=0.47\textwidth]{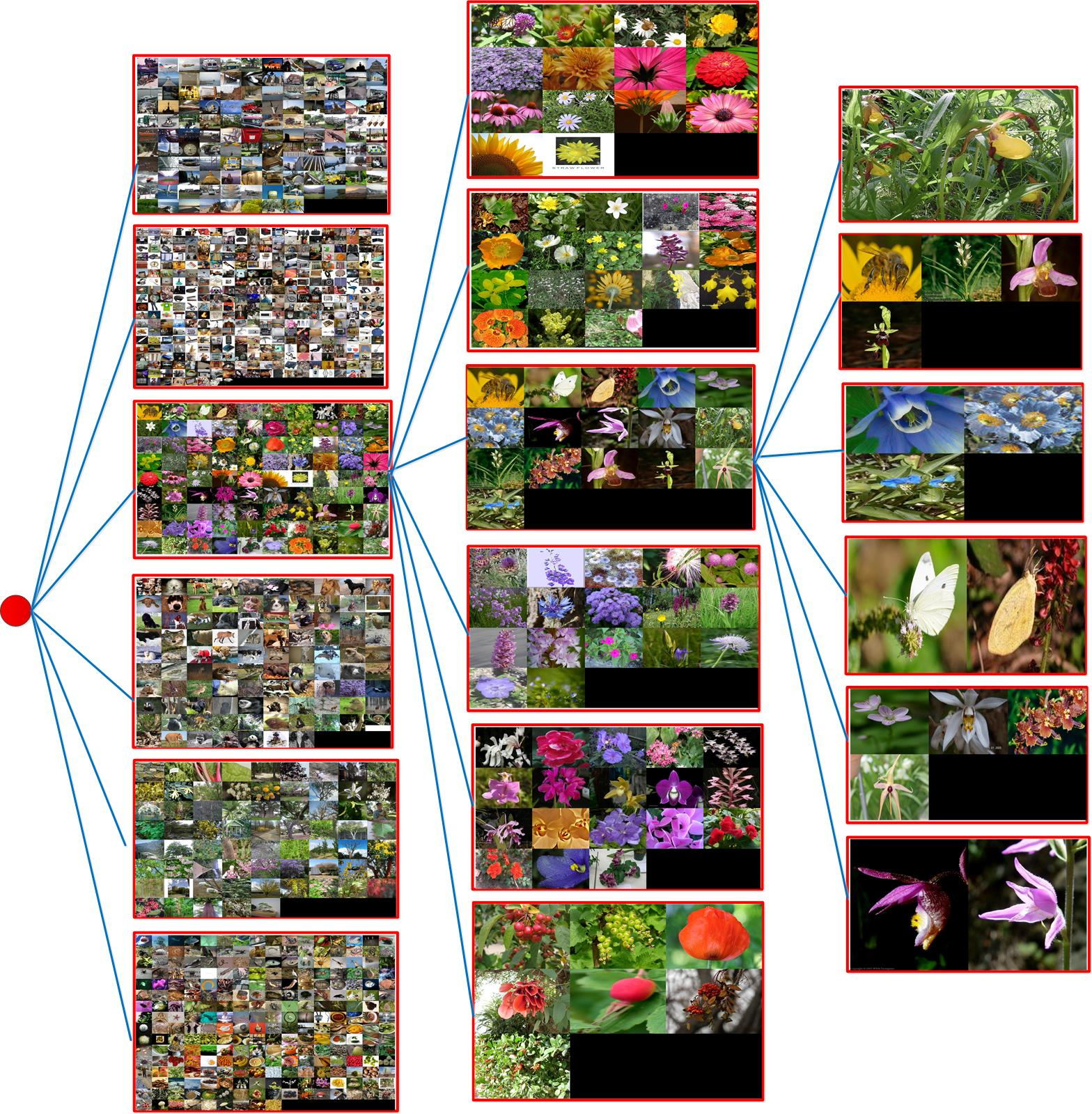}
\caption{Part of the visual tree $T_{6,4}$ based on CNN features for ILSVR2010.}
\label{fig:visualTree2}
\end{figure}

The parameters $K$ and $L$ are predefined. Hereafter, we denote the visual tree of depth $L$ with branching factor $K$ by $T_{K,L}$. The pseudo-code of visual tree construction is presented in Algorithm \ref{algorithm 1}. Note that the set of classes contained in the node $v$ is denoted by $C(v)$. The $i$th children node of the node $v$ contains the class set $C_{v}^{i}$. For each node $v$, the union of the class sets contained in its child nodes is equal to the class set contained in the node $v$, that is, $\bigcup\limits_{i\in C(v)}C_{v}^{i}=C(v)$. Moreover, any pairwise sibling nodes do not overlap, which satisfies $C_{v}^{i}\cap C_{v}^{j}=\emptyset$, $i\neq j$. The clustering results for ILSVRC2010 are visualized in a visual tree $T_{6,4}$  based on CNN features in \figurename \ref{fig:visualTree2}, where the membership between categories can be clearly observed. Similar classes are clearly clustered coarsely into a group.

\textbf{Computational complexity comparison.} We next analyze the cost of constructing the hierarchical category structure. The cost of spectral learning on $N$ classes is known to be $O(N^3)$. With $m$ examples, $N$ classes, and $D$-dimensional features, the affinity matrix cost is $O(2NmD + N^2D)$, which  is much smaller than the cost of tree construction based on confusion matrices \cite{5item,10item,16item}. For example, the label tree \cite{5item} is built based on a confusion matrix. To obtain a  confusion matrix, $N$ one-vs.-all SVM classifiers need to be trained. The element of the confusion matrix at position $(i,j)$ is the number of samples of the $i$-th class that are classified as the $j$-th class label. The cost of training an SVM classifier is between  $O(N_{sv}^3 + mN_{sv}^2 + mDN_{sv}))$ and $O(Dm^2)$, where $N_{sv}$ is the number of support vectors and $m$ is the number of training data. Thus, the total cost of confusion matrix construction by training $N$ SVM classifiers is between  $O(mDN + \sum_{i=1}^{N}(N_{svi}^3 + mN_{svi}^2 + mDN_{svi}))$ and $O(NDm^2)$, where $N_{svi}$ is the number of support vectors for the $i$-th classifier. In general, because $m\gg N$, the cost of tree construction based on confusion matrices is much greater than that of the clustering-based tree construction. Thus, clustering-based tree construction has computational advantages over confusion-based tree construction.

\subsection{N-best path for hierarchical Learning}
\begin{figure}[!t]
\centering
\includegraphics[width=0.47\textwidth]{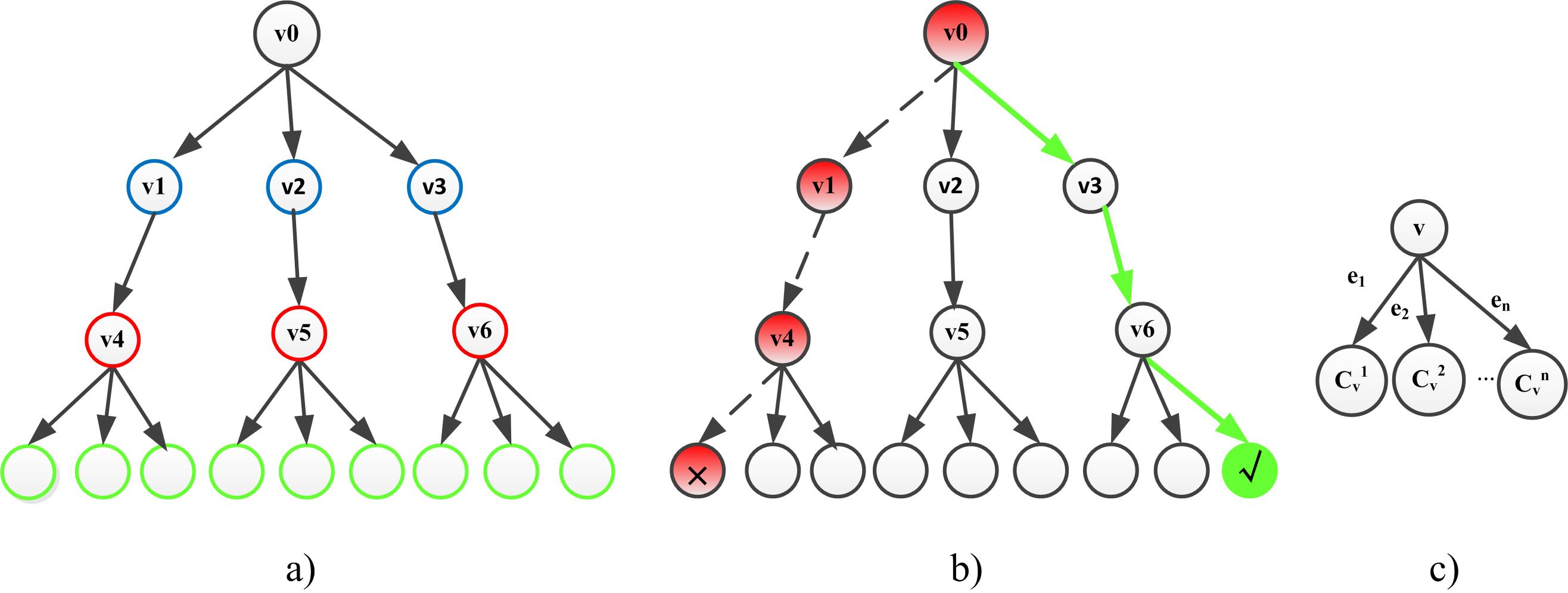}
\caption{The tree-based model for class prediction. a) A visual tree model. b) Error propagation in greedy learning. c) The node notation.}
\label{fig:treeModel3}
\end{figure}

\textbf{Visual tree model.} Given a visual tree, a tree-based model for class prediction can be constructed. An example of a visual tree is shown in \figurename \ref{fig:treeModel3}a. Let us denote the tree-shaped hierarchy as $T=\{V,E\}$, where $V$ is a set of nodes and  $E$ a set of edges. Each node $v$ contains no more than $K$ children nodes. In the following, we also use the symbol $C_{v}^{i}$ denote the $i$th child node of the node $v$. Each edge $e_{vi}$ is associated with a classifier from $W_v = \{w_v^i \in \mathbb{R}^D\}_{i=1}^{K}$ with its score function defined as
\begin{equation}
\label{equ 8}
S_{v}^{i}(x)=x^{T}w_{v}^{i}
\end{equation}
Here, $x\in \mathbb{R}^{D}$ is an input vector. Obviously, the set of classes at the $i$-th child of node $v$, $C_{v}^{i}$, is a subset of $C(v)$, which is shown in \figurename \ref{fig:treeModel3}c.

\textbf{Learning criterion.} Given a tree-based model, the greedy algorithm is typically applied to predict the class of an input image $x\in \mathbb{R}^{D}$, where a single path is explored from the root node to the leaf node. From the root node, an edge corresponding to the largest classification score is selected, e.g., $e=\argmax\limits_{j}S_{v}^{j}(x)$. The same selection operation is iteratively applied to traversing child nodes in the subsequent layers until a leaf node is reached. Thus, class prediction is transformed into a path exploration that is a concatenation of the edges,  $[e_{1},e_{2},\cdots,e_{L}]$, where $L$ is the path length. This implies that the greedy algorithm only keeps one node in each layer. The best computational complexity is $log_{K}N$, where $N$ is the number of the classes and $K$ is the number of branches per node. However, one disadvantage of the greedy algorithm is that an error made at a higher layer in the tree-based model cannot be corrected. As shown in \figurename \ref{fig:treeModel3}b, the correct node is the node $v6$, but the greedy algorithm makes the decision using the node $v4$. That is, once the greedy algorithm makes a wrong decision at a higher level, it proceeds in the wrong direction without an opportunity to correct the mistake.

Therefore, we adopt the best path algorithm to avoid error propagation. We transform the class prediction problem to a path-searching problem. The scoring function is defined as the joint probability for a path $\emph{P}=[e_{1},e_{2},\cdots,e_{L}]$. Suppose that the tree hierarchy is subject to a causal Bayes network and the child node is only conditionally dependent on its parent node and is independent of its ancestors, then a probability that an input image $x$ goes through the edge $e$ is defined as
\begin{equation}\label{equ 5}
    p(e|v) = \frac {1}{1+exp(-S_v^e(x))}
\end{equation}
where $S_v^e(x)$ is the edge score function obtained from (\ref{equ 8}). For a query image, it should traverse the entire visual tree and find a path with the maximum joint probability. We formulate the problem as
\begin{equation}\label{equ 6}
    \emph{P}^{*}=\argmax\limits_{\emph{P}}p(\emph{P})=\argmax\limits_{\emph{P}}p(v_0)\prod\limits_{i}p(e_i|v_i)
\end{equation}
where \emph{P} is a path.

Since the best path algorithm computes all edge scores for all the nodes in a layer, the computational complexity is a little higher. We adopt the approximate dynamic programming algorithm to find the maximum confidence path. Specifically, if we want to achieve a path with the maximum joint probability at the $i$-th layer, we should achieve the maximum probability at the $(i-1)$-th layer. To reduce the computational complexity, we only keep the first $Q$ best paths. We formulate the problem as
\begin{equation}\label{equ 7}
    \max\limits_{P}p(P_{t+1}^{i})=\max\limits_{e_{t+1}}p(e_{t+1}|v_{t})\max\limits_{P_{t}}p(P_{t}^{i})
\end{equation}
where $i = 1,..., Q$.
In each layer, we keep the first $Q$ best branches corresponding to the first $Q$ largest probabilities. We name the approximation method searching for the optimal path \emph{N-best path}.

\textbf{Algorithm description.} Our algorithm is detailed in Algorithms \ref{algorithm 2} and \ref{algorithm 3}. Algorithm \ref{algorithm 2} shows how to compute the scoring function for each edge in the tree. Algorithm \ref{algorithm 3} shows how to predict the class of a query image according to the tree model. An edge corresponds to a classifier. In detail, in Algorithm \ref{algorithm 2}, the edges linking with the same parent nodes are trained simultaneously. That is, we take a one-vs.-all SVM to train a classifier. A parent node $v$ has no more than $K$ edges linking with the child nodes $\{C_{v}^{1},C_{v}^{2},\cdots \}$, where $C_{v}^{i}$ contains a set of image classes. For the edge $e_{vi}$, we train a classifier formulated as in (\ref{equ 8}) as the score function. Specifically, the positive samples are from node $C_{v}^{i}$, and the negative samples are from the other sibling nodes $\{C_{v}^{l}\}_{ l\neq i}$. We train classifiers depending on the tree hierarchy. That is, a classifier for a node is only related to the set of classes contained in the same parent node in the tree model. Our algorithm achieves a more accurate classifier because it partly avoids the imbalance between the positive and negative samples.

Algorithm \ref{algorithm 3} describes our algorithm for predicting the input image class. We define a path $\emph{P}=[v;e_{1},\cdots,e_{L}]$ as a route with increasing levels in the hierarchy from node $v$ to a leaf node following a sequence of selected edges $[e_{1},\cdots,e_{L}]$. We also define a ¡°branch¡± $ \emph{P}(v)=[v;e_{1},\cdots,e_{t})$ as a part of path $\emph{P}$  that overlaps $\emph{P}$ with length $t$. $\emph{P}(v)$ contains several possible routes to the leaf nodes, and the path  $P$ represents only one of them. To begin the search, we initialize the branch $\emph{P}(v)=[0;)$, which corresponds to the set of paths in the whole tree. We then split the branch into $K$ sub-branches, where $K$ is a branching factor, $\{[P(v),e)\}_{e}$. We compute the edge weight and the joint probability of the current branch. In each layer, we keep the first $Q$ best branches corresponding to the first $Q$ largest probabilities.

\begin{table}[!t]
\renewcommand{\arraystretch}{1.3}
\caption{The computational complexity of our approach}
\label{table 1}
\begin{center}
\begin{tabular}{lcl}
\hline
\hline
                        &   $Cost$  & Unit            \\  \hline
Tree construction       &   $O(2DNm + N^2D)$  &  Multiplication  \\  \hline
Clustering              &   $O(N^3)$   & Multiplication         \\  \hline
Traversing edges  &   $O(KQlog_{K}N)$& Classifier number  \\  \hline
Path probability                   &   $O(Q\log_{K}N)$ & Multiplication  \\  \hline
\end{tabular}
\end{center}
\end{table}

\textbf{Computational complexity analysis.} Class prediction consists of two parts: the edge weight and the path probability. The edge weight is related to the classifier, so we only count the number of classifiers as the computational complexity. For the computation of path probability, we simply count the number of the multiplication. A queried image should traverse the entire visual tree. Thus, in the computation of (\ref{equ 6}), the number of classifiers is $\sum_{i=1}^{L}K^{i}\approx K^{L}$ and the multiplication number is $(L-1)*K^{L}$ for the computation of the path probability. This is obviously time consuming. However, the \emph{N-best path} algorithm in (\ref{equ 7}) does not compute the whole edge-scoring function, thereby reducing the computational complexity to $K+(L-1)QK \approx LQK$ classifiers and $(L-1)Q$ multiplications for the path probability, which is much lower than the traditional best path algorithm. The proposed algorithm always carries out the search over the $Q$ best candidate branches. The search terminates when the branch contains only one path; that is, it reaches a leaf node. Table \ref{table 1} shows the main cost at each intermediate step.

Furthermore, we extend our model using ensemble decisions and aim to improve image classification performance by using multiple visual trees. We randomly divide the training data into five sets. According to our basic framework, five visual trees are obtained independently.  Five visual trees are used to infer an input query image, producing five results. The weighted average of five results in each class is computed, and the final decision depends on the final confidence score. The ensemble decision can improve classification accuracy, which is proved in the experiments below.

\begin{algorithm}
\caption{Training a tree model}
\label{algorithm 2}
\begin{algorithmic}

\REQUIRE {The indexes of the classes in each node, the samples of each class}
\ENSURE {Scoring function of each edge}

\FOR{layer $v=0$ \TO $L-1$}
    \FOR{node $C_{v}^{i}$ in the layer $v$ \AND $C_{v}^{i}$  is not a leaf node}
            \FOR{edge $e_{vi}$, which links the node $v$ with the node $C_{v}^{i}$}
                \STATE Construct the positive set containing the samples from the node $C_{v}^{i}$
                \STATE Construct the negative set containing the samples from the other sibling nodes $C_{v}^{j}, j \neq i$
               \STATE Train an SVM classifier for edge $e_{vi}$
            \ENDFOR
    \ENDFOR
\ENDFOR

\end{algorithmic}
\end{algorithm}

\begin{algorithm}
\caption{Inference of the class prediction}
\label{algorithm 3}
\begin{algorithmic}

\REQUIRE {The tree model, a query image}
\ENSURE {The class prediction}

\STATE Set node index $v \leftarrow 0$ \COMMENT{Starting from root node branch}
\STATE Set branch $\hat{\emph{P}} \leftarrow P(0)$
\STATE Set priority queue $Q \leftarrow \emptyset $
\STATE Set scoring function $S(x) \leftarrow 1$
\REPEAT
    \FOR{each branch $\hat{\emph{P}}\in Q$}
        \STATE Split $\hat{\emph{P}}$ into $[\hat{P},1),\cdots,[\hat{P},K) $ according to the tree structure
        \FOR{edge $e_{i}$}
            \STATE Compute the score $p(e_{i}|v)$ according to (\ref{equ 5})
            \STATE Update accumulated joint probability $S(x)=p(e_{i}|v)S(x)$
        \ENDFOR
    \ENDFOR
    \STATE Push the top $N$ largest score paths $(v,[\hat{\emph{P}},e),S(x))$ into $Q$
\UNTIL{$\mid \hat{\emph{P}} \mid\neq 1$}
\STATE Retrieve the largest score path $(v,[\hat{\emph{P}},e),S(x))$ in $Q$
\STATE $\emph{P}^{*} \leftarrow [\hat{\emph{P}}]$
\RETURN{The class label $e_{i}$}

\end{algorithmic}
\end{algorithm}


\section{Experimental Results}\label{sec:experimental results}
We test our method on two challenging image datasets: ILSVRC2010 and Caltech 256, the most popular image datasets for image classification. ILSVRC2010 has three parts: 1) a training set including $1.2$ million training images in $1000$ image categories (the number of images per category varies from $668$ to $3047$); 2) a verification set containing 50k images with $50$ images per category; and 3) a test set of 150k images with $150$ images per category. Since ILSVRC2010 contains a test set but other ILSVRC datasets do not, we evaluate our approach on ILSVRC2010. Caltech-256 consists of $256$ object categories with $30607$ images and a background category. Each category contains at least $80$ images.

The primary aim of our experiments is the evaluation of classification performance. Furthermore, we investigate the influence of some critical factors on clustering and classification performance, such as the deep features learned by the deep learning network, the similarity metric used in clustering, and the visual tree model structure. As an example, we only give the results on $T_{32,2}$, $T_{10,3}$, $T_{6,4}$. To compare our approach with state-of-the-art methods, we use the top-$1$ and top-$5$ classification accuracy as the criteria on ILSVRC2010 and the classification accuracy as the criterion on Caltech 256; these have been extensively used as evaluation criteria on the two datasets. In our experiments, a linear SVM based on the LIBLINEAR toolbox \cite{31item} is used to train SVM classifiers. All the experiments are done on DELL Precision T7500 with a Geoforce TITANX GPU with $12$ GB memroy.

We first introduce the experimental setup. Given a visual tree, we first detail how an edge is associated with a classifier. For an edge in the first layer whose parent node is the root node, and taking computational efficiency into account, we simply randomly sample $600$ images from each class training dataset as the new training samples. For an edge in other layers, all the samples contained in the training set are used. For an edge $e_{vi}$, we train an associated classifier on a newly constructed training dataset. Specifically, the positive samples are from all the classes contained in the child node $C_v^i$ linked with edge $e_{vi}$, and the negative samples are from all the classes contained in the sibling child nodes, $C_{v}^{j}, j\neq i$, which have the same parent node as the child node $C_{v}^{i}$. A linear SVM classifier is then learnt from the training data and associated with an edge. During label inference, we only keep the top $5$ nodes corresponding to the first $5$ largest probabilities.

We use two pre-trained deep learning models trained on ILSVRC2012 to represent an image: Inception V3 \cite{46item} and AlexNet \cite{29item}. For the two deep learning models, we use the output of the second to last layer as the feature vector. The CNN feature is $4096$-dimensional and the Inception feature is $2048$-dimensional. AlexNet contains $8$ layers and Inception V3 contains $42$ layers.

We consider two important components: visual tree construction and class prediction. We construct the following variations of our approach with different component combinations:
\begin{enumerate}
    \item Single Tree + Best Path decision (ST-BP). $T_{32,2}$ is used as a single visual tree model and the \emph{N-best path} decision is implemented for class prediction.
    \item Single Tree + Greedy decision (ST-G). $T_{32,2}$ is used as a single visual tree model and the greedy learning is implemented for class prediction.
    \item Single Tree + Best Path decision + Multiple prediction (ST-BP-M). In this variation, the visual model and class prediction method are similar to the first variation. The difference is in the number of predictions: for a query image, we not only estimate the original query image but also estimate the five image crops from the four corners and the center of the original image as in CNN \cite{28item}.
    \item 5 Trees + Best Path decision + Multiple prediction (5T-BP-M). This method is similar to the third variation, the only difference being the number of trees. This variation uses five visual trees to infer the class prediction.
\end{enumerate}

\subsection{Comparison with state-of-the-art hierarchical learning methods}
 We compare our approach with four state-of-the-art hierarchical learning methods which have experimental results on ILSVRC2010: the label tree classifiers \cite{5item}, the fast label tree classifiers \cite{16item}, the probabilistic tree classifiers \cite{10item}, and the hierarchical cost sensitive classifiers \cite{30item} in terms of classification accuracy. The results are shown in Table \ref{table 3}. Our visual tree achieves the best classification accuracy. Furthermore, deep features achieve better classification than the other (crafted) features. Comparing the two types of deep features, the Inception feature achieves $76.2\%$ classification accuracy, which is $15\%$ greater than that of the CNN feature; the Inception feature appears to be more discriminant than the CNN feature. Thus, it implies that the greater the number of layers in the deep learning network, the more discriminant the feature. Note that over $30\%$ and $40\%$ absolute performance gains are achieved using our approach with CNN and Inception V3, respectively, compared to the next best method.


\begin{table}[!t]
\renewcommand{\arraystretch}{1.3}
\caption{Classification accuracy (\%) comparison on ILSVRC2010 with different tree structures}
\label{table 3}
\begin{center}
\begin{tabular}{lccc}
\hline
                            &   $T_{32,2}$  &   $T_{10,3}$      &   $T_{6,4}$   \\ \hline
Inception3 ST-BP            &   76.2        & 74.6              & 73.6          \\ \hline
CNN ST-BP                   &   61.2       &   60.4           &   58.7       \\ \hline
Hierarchical cost sensitive
classifiers\cite{30item}    &   28.3       &	26.7           &   24.2       \\ \hline
Probabilistic tree
classifiers\cite{10item}    &   21.4       &	20.5           &	17.0       \\ \hline
Label tree classifiers
\cite{5item}                &   8.3        &	6.0            &	5.9        \\ \hline
Fast label tree
classifiers\cite{16item}    &   11.9       &	8.9            &	5.6        \\
\hline
\end{tabular}
\end{center}
\end{table}

\begin{figure}[!t]
\centering
\includegraphics[width=0.47\textwidth]{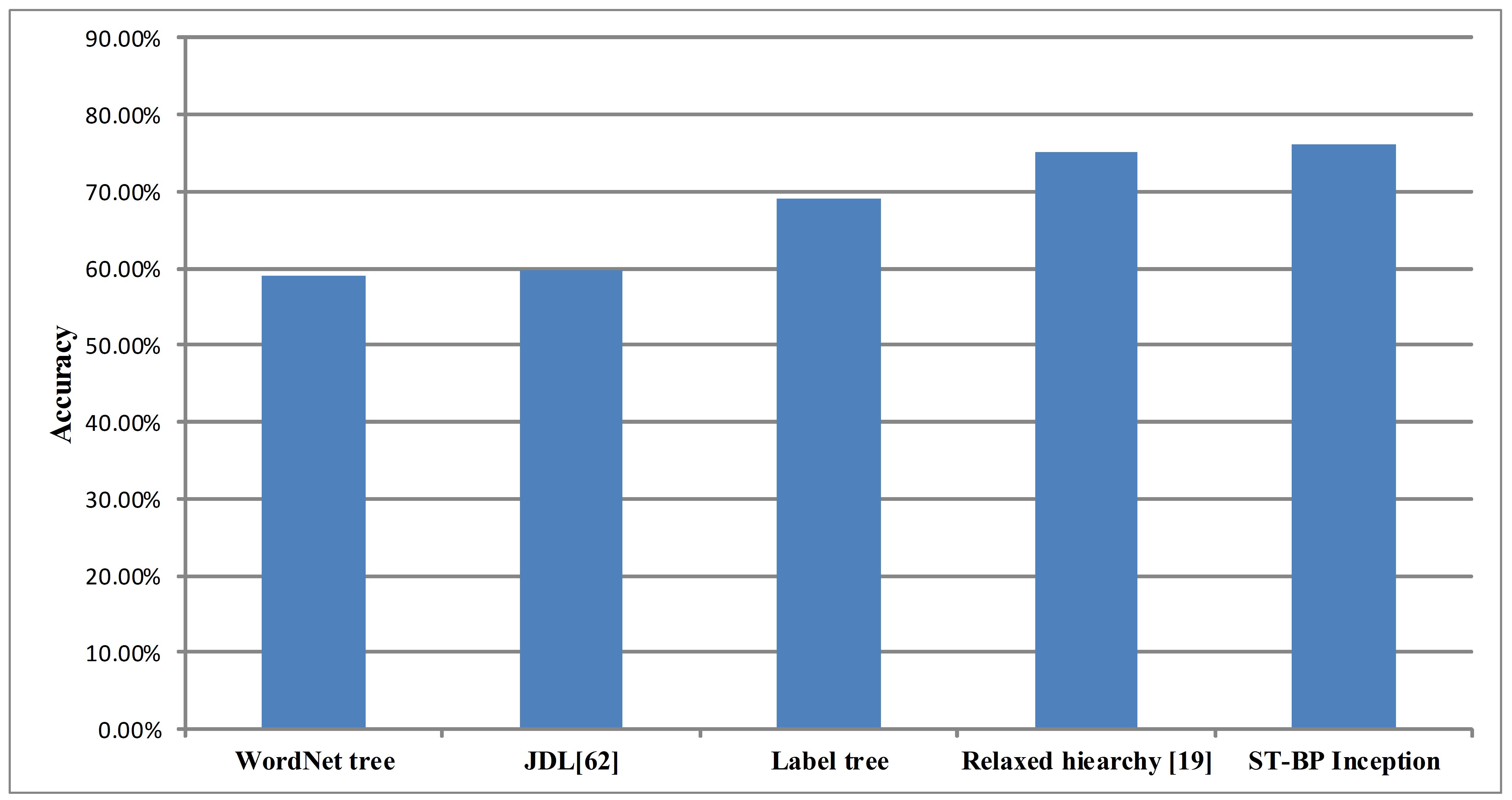}
\caption{Comparison of different hierarchical methods on the Inception feature.}
\label{fig:comparisonHiearchy}
\end{figure}

 The multi-class classification performance is greatly influenced by two important factors: feature representation and hierarchical learning. As shown in Table \ref{table 3}, the compared hierarchical methods used very different features with each other. The more distinctive the feature is, the better the classification performance is. In order to investigate the effect of the hierarchical learning, we compare the following hierarchical methods under the same feature: the WordNet tree, the label tree, JDL\cite{6item}, Relaxed hiearchy \cite{7item}  and our method. They represent five typical hierarchical learning methods.

WordNet is a semantic structure according to taxonomy which is presented on the ImageNet website. We construct a two-layer tree according to the taxonomy distribution of 1000 categories of ILSVRC2010. The first-layer nodes are from the first-level nodes of WordNet which contains the 1000 categories of ILSVRC2010, and all the categories are treated as the leaf nodes whose hierarchical relations to the first-layer nodes agree with the taxonomy. Note that not all the categories of ILSVRC2010 are the leaf nodes in WordNet and it is an unbalanced tree. The WordNet tree is denoted as $T_{7,2}$ whose branch factor is seven.

Label tree based methods are an important branch of hierarchical learning methods. Probabilistic tree\cite{10item}, Label tree\cite{5item} and Fast label tree\cite{16item} all belong to Label tree based method. We build a simple label tree $T_{32,2}$ and predict a query image based on a greedy learning scheme.

JDL\cite{6item} is the latest visual tree constructed based on AP clustering. We build a visual tree $T_{32,2}$ using the Inception feature. JDL is different from other compared methods because each middle node has different feature representation which is computed by joint dictionary learning.

Relaxed hiearchy \cite{7item} allows each node to neglect the confusing classes. In other words, a class can be contained in more than one node. This method is only suitable for the moderate dataset, which is implemented on Caltech $256$ and SUN $397$ in \cite{7item}. When the number of categories becomes larger, the categories become more confused with each other. And the nodes will grow exponentially which results in prohibited computations. In our experiments, we just construct a shallow binary tree $T_{2,3}$ using the source code presented by the authors.

 \figurename \ref{fig:comparisonHiearchy} shows the comparison results. Under the conditions of the Inception feature, our method still achieves the best result among different hierarchical methods. The WordNet tree is inferior to our method, which implies that there is a gap between the taxonomy and the visual classification. Thus, the taxonomy structure is not suitable for multi-class visual classification. Moreover, Label tree and Relaxed hierarchy which used confusing matrix to build a tree are more time consuming than our method let along their classification accuracies are lower than our method. Our method is superior to JDL\cite{6item}, and the accuracy difference is $16.43\%$, because their greedy learning based prediction cannot avoid the error propagation.


\subsection{Comparison with representative state-of-the-art models on ILSVRC2010}
 We next compare our method with seven representative image classification methods: HOG+LBP+sparse coding [34], SIFT +Fisher vector \cite{33item}, Fisher vector \cite{34item}, one-vs.-all SVM, JDL \cite{6item}, the hierarchical tree cost sensitivity classifier \cite{30item}, and the hierarchical tree structure SVM classifier \cite{4item}. The first four methods represent the flat classification mechanism, and the last three methods are hierarchical. Moreover, \cite{6item} and \cite{30item} are greedy learning methods, and \cite{4item} and our approach are based on the optimal path searching solution. However, \cite{4item} utilizes the structured SVM to solve the class prediction problem, while our method implements the best path search to obtain the solution. The one-vs.-all SVM is a popular method for multi-class classification. We also treat the one-vs.-all SVM combined with the Inception feature as a benchmark method. A query image is designated to be in a class with the maximum confidence value. The results are shown in Table \ref{table 4}. All the results presented for competing methods are the original published results; in \cite{6item} and \cite{30item}, the authors did not calculate the top-5 results, so these are denoted by "--" (not available). It proves again that the methods using deep features outperform the other methods using traditional features, demonstrating that deep features are more discriminative than the traditional features. Furthermore, with the same deep feature, ST-BP achieve better classification accuracy than One-vs.-all+Inception by $1.5\%$. It implies that our method can remarkably improve the classification performance for imbalanced data. In view of the class prediction method, ST-BP is superior to ST-G on CNN features, with an absolute gain of over $5\%$.

\begin{table}[!t]
\renewcommand{\arraystretch}{1.3}
\caption{The comparison of classification accuracy (\%) on ILSVRC2010}
\label{table 4}
\begin{center}
\begin{tabular}{@{}p{2.5cm}lccccc}
\hline
\textbf{Method}                             &   \textbf{Top-1}  &   \textbf{Top-5}  &   \textbf{Flat}   &   \textbf{Greedy}     &   \textbf{Path}   \\ \hline
Signature + Fisher Vector \cite{33item}     &   54.3            &   74.3            &   $\surd$         &       --              &       --          \\ \hline
Fisher Vector \cite{34item}                 &   45.7            &	65.9            &   $\surd$         &       --              &       --          \\ \hline
HOG + LBP + CODING \cite{55item}            &   52.9            &	71.8            &   $\surd$         &       --              &       --          \\ \hline
JDL+AP Clustering  \cite{6item}             &   38.9            &	--              &      --           &     $\surd$           &       --          \\ \hline
Hierarchical cost sensitivity
classifier \cite{30item}                    &   41.1            &	--              &      --           &     $\surd$           &       --          \\ \hline
Structured SVM \cite{4item}                  &   23.0            &   --              &      --           &       --              &     $\surd$       \\ \hline
 One-vs.-all+Inception                     &   74.7           &   90.5             &   $\surd$         &       --              &       --          \\ \hline
CNN ST-G                                    &   56.1            &   --              &     --            &                        &
\\ \hline
CNN ST-BP                                   &	61.2            &	81.7            &      --           &       --              &     $\surd$
\\ \hline
Inception3 ST-BP                       &	76.2            &	91.1            &  --       &       --              &      $\surd$           \\ \hline
\end{tabular}
\end{center}
\end{table}

\subsection{Comparison with deep learning network }
It is worth noting that CNN \cite{28item} was a major milestone in image classification, with many deep learning networks developed thereafter. Inception V3 \cite{46item} is one of the latest versions issued by Google Inc. We compare our method with CNN \cite{28item}. CNN's  success can be attributed to many factors that include a tuned architecture, augmented training data, and ensemble decision-making. Inspired by CNN \cite{28item}, we pay particular attention to the last two factors. In \cite{28item}, $4$ corner patches and $1$ center patch were cropped and then flipped, so $10$ images were added to the training data. In our method, we only crop 5 image patches as in \cite{28item}. As shown in Table 4, our method achieves comparable results to CNN \cite{28item}. The difference between CNN1 and CNN2 is that CNN1 makes 10 predictions for a query image while CNN2 only makes a single prediction for a query image. Here, we cite the CNN results given in \cite{28item}. Considering the similar decision rule, we compare ST-BP with CNN2. The top-1 score of our method is higher than CNN2, while their top-5 scores are similar. With respect to multiple predictions, we use the entire image and its five crops for class prediction. Comparing ST-BP-M with CNN1, even though the augmented training set used in our approach is smaller than in CNN, we achieve comparable top-1 and top-5 scores to CNN1, and ST-BP-M is superior to CNN2. Comparing 5T-BP-M with CNN1, both use ensemble decision-making, but the size of our augmented data and the number of ensemble predictions are smaller than those of CNN1. The results demonstrate that our method can slightly outperform CNN when they are set in a similar environment. Augmented data and ensemble decision-making can improve image classification performance. Our method is simpler than CNN \cite{28item}, because the number of parameters required for our method is much smaller. We also unify the latest Inception feature using our model. Table \ref{table 5} demonstrates that the Inception feature achieves the best results compared to other methods and is more discriminative than the CNN feature. For both single and multiple visual trees, the performance of the Inception feature is higher by $(15\%, 9.4\%)$ and $(16.5\%, 10\%)$ in terms of top-1 and top-5 results, respectively, than the CNN feature. It can also be seen that multiple predictions provide even greater gains than ensemble learning. For a single tree with CNN, multiple predictions achieve $0.8\%$ and $0.7\%$ improvements in terms of top-1 and top-5 results, respectively. For multiple trees with Inception features, multiple predictions achieve $2.7\%$ and $1.6\%$ improvements in terms of top-1 and top-5 results, respectively.

\begin{table}[!t]
\renewcommand{\arraystretch}{1.3}
\caption{Comparison With CNN \cite{28item} on ILSVRC2010}
\label{table 5}
\begin{center}
\begin{tabular}{@{}ccccc}
\hline
\textbf{Method} & \textbf{Top-1} & \textbf{Top-5} & \textbf{Data Augment} & \textbf{Prediction} \\ \hline
 CNN1       & 62.5	 & 83.0 & 5 crops + flip& 10 \\ \hline
 CNN2       & 61.0  & 81.7 & 5 crops + flip & 1  \\ \hline
 CNN ST-BP   & 61.2	 & 81.7 & no             & 1  \\ \hline
CNN ST-BP-M  & 62.0  & 82.4 & no             & 6  \\ \hline
CNN 5T-BP-M  & 62.4  & 82.7 & no             & 6  \\ \hline
Inception3 ST-BP & 76.2 & 91.1 & no          & 1  \\ \hline
Inception3 ST-BP-M &  77.4   & 91.8   & no & 6 \\ \hline
Inception3 5T-BP & 76.2   & 90.8   & 5 crops  & 1 \\ \hline
Inception3 5T-BP-M & 78.9    & 92.4     & 5 crops  & 6 \\
\hline
\end{tabular}
\end{center}
\end{table}

\begin{figure}[!t]
\centering
\includegraphics[width=0.47\textwidth]{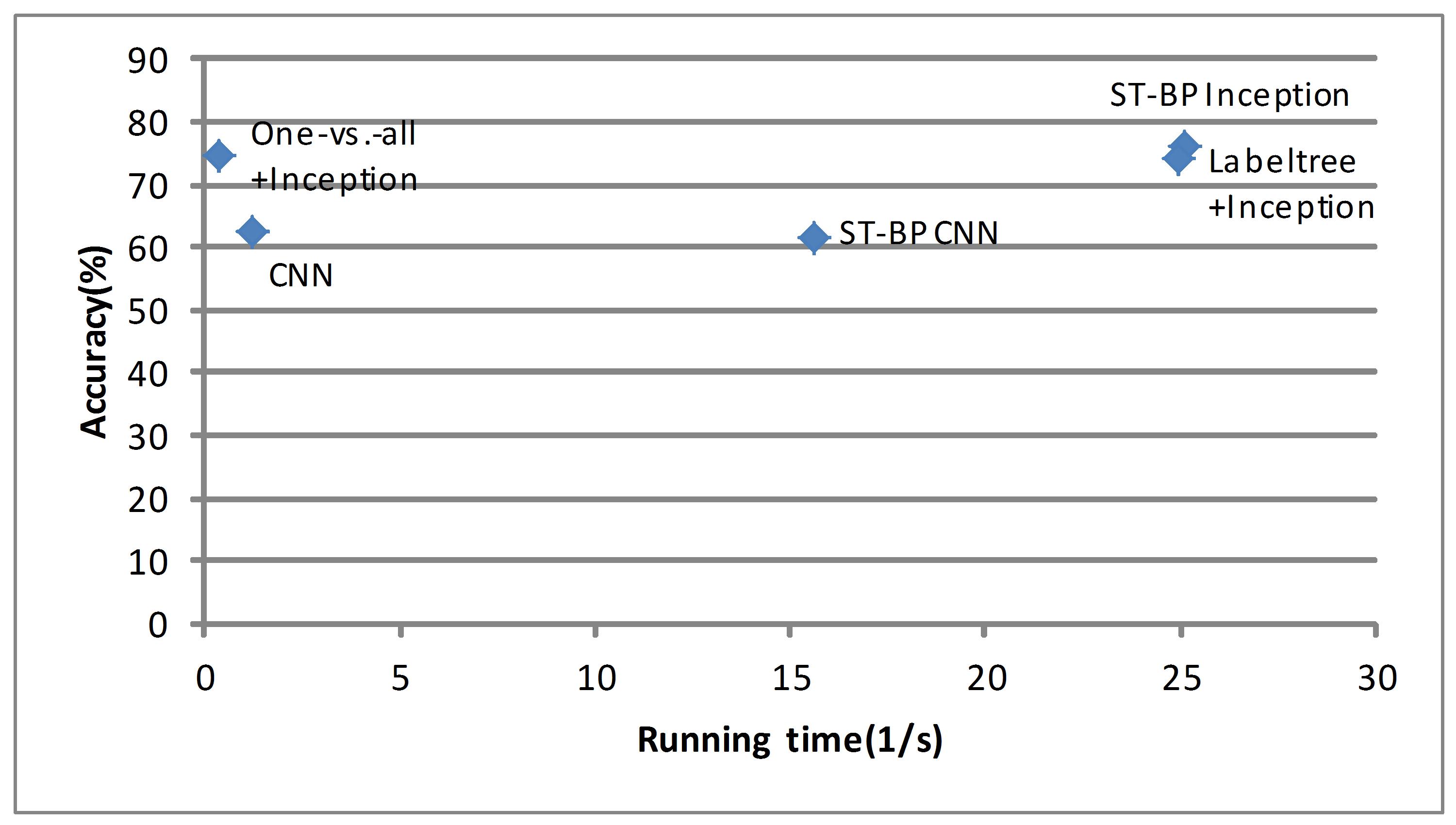}
\caption{Running time vs. classification accuracy.}
\label{fig:efficient_comparison}
\end{figure}

\begin{figure}[!t]
\centering
\includegraphics[width=0.47\textwidth]{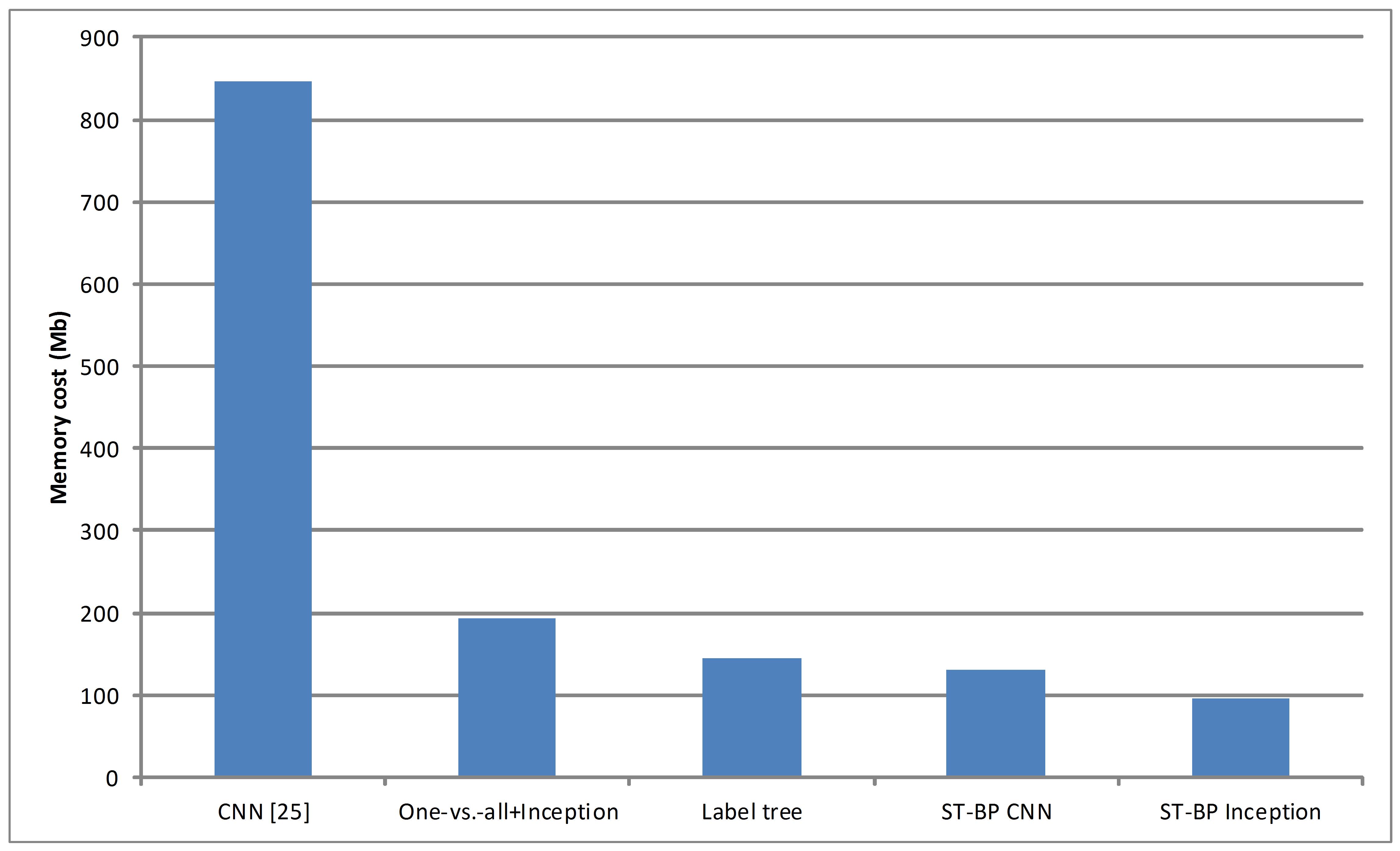}
\caption{Comparison on average CPU memory cost for a query image.}
\label{fig:memory_comparison}
\end{figure}

 Furthermore, we discuss the running time and the CPU memory cost when a query image is tested. We compare the following methods: one-vs.-all SVM, CNN \cite{28item}, Label tree with the Inception feature, ST-BP CNN and ST-BP Inception. The relation between classification accuracy and the average running time for a query image is shown in \figurename\ref{fig:efficient_comparison}. The comparison of CPU memory cost is presented in \figurename\ref{fig:memory_comparison}.
Comparing CNN\cite{28item} with ST-BP CNN and ST-BP Inception, the latter is faster than CNN, because CNN\cite{28item} has to spend more time on loading the model to the CPU memory while our method, ST-BP CNN and ST-BP Inception, does not load all the SVM models at a time. For each node of the visual tree, we call the required SVM models from the disk. Moreover, the CPU memory cost of CNN \cite{28item} is $8.8$ times as much as the one of ST-BP Inception and is $6.5$ times as much as the one of ST-BP CNN.
Label tree combined with the Inception feature is similar to ST-BP CNN in the running time and CPU memory cost, but the classification accuracy is lower than ours.
The one-vs.-all SVM is a typical flat classification method. With the same feature, its classification accuracy is smaller than ours by $1.5\%$ and the running time is much longer than ours in  \figurename\ref{fig:efficient_comparison}. From \figurename\ref{fig:memory_comparison}, the CPU memory cost is about 2 times as much as ours because it requires to load all $1000$ SVMs to the CPU memory. The experimental results of comparison between the one-vs.-all SVM and our method agree with the comparing analysis of computational complexity in Section III.
To sum up, our method has the distinct advantage of the computational complexity, besides it achieves the comparative results on classification accuracy to the latest deep learning method.

\subsection{Interpretation of clustering results}
\begin{figure}[!t]
\centering
\includegraphics[width=3.2in]{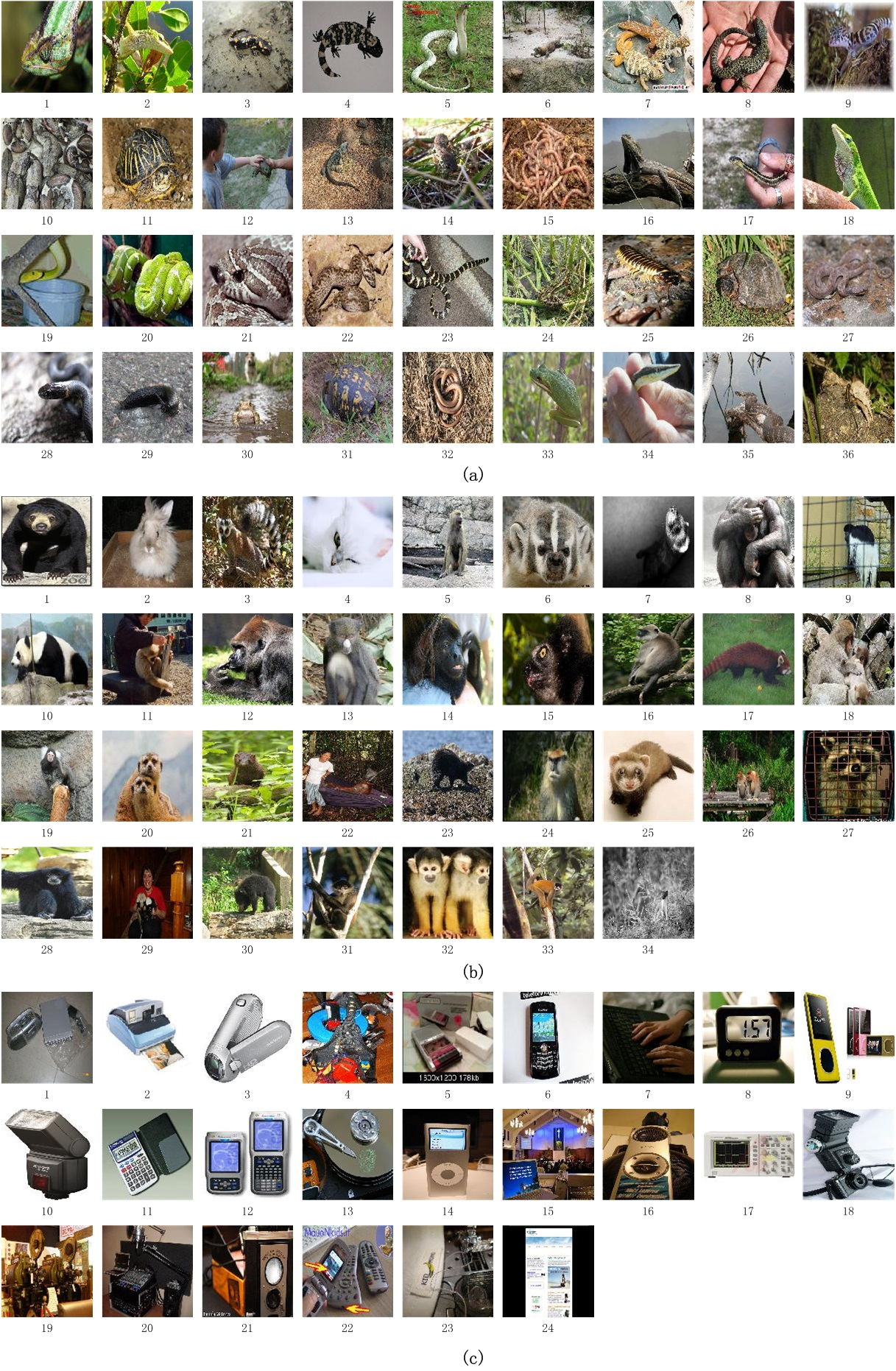}
\caption{Three groups clustered using our approach with $T_{32,2}$ on ILSVRC2010.}
\label{fig:threeGroups6}
\end{figure}

\begin{table*}[t!]
\centering
\renewcommand{\arraystretch}{1.3}
\caption{The corresponding class names to Fig. \ref{fig:threeGroups6}. }
\label{table 9}
\tabcolsep=8pt
\begin{minipage}{20cm}
\footnotesize
 \begin{tabular}{p{0.5cm}|p{16cm}}
  \hline
 Group  & Class name \\ \hline
 a) & 1. African chameleon, Chamaeleo chamaeleon
2. American chameleon, anole, Anolis carolinensis
3. European fire salamander, Salamandra salamandra
4. Gila monster, Heloderma suspectum
5. Indian cobra, Naja naja
6. Komodo dragon, Komodo lizard, dragon lizard, giant lizard, Varanus komodoensis
7. agama
8. alligator lizard
9. banded gecko
10. boa constrictor, Constrictor constrictor
11. box turtle, box tortoise
12. bullfrog, Rana catesbeiana
13. common iguana, iguana, Iguana iguana
14. common newt, Triturus vulgaris
15. earthworm, angleworm, fishworm, fishing worm, wiggler, nightwalker, nightcrawler, crawler, dew worm, red worm
16. frilled lizard, Chlamydosaurus kingi
17. garter snake, grass snake
18. green lizard, Lacerta viridis
19. green mamba
20. green snake, grass snake
21. hognose snake, puff adder, sand viper
22. horned viper, cerastes, sand viper, horned asp, Cerastes cornutus
23. king snake, kingsnake
24. leopard frog, spring frog, Rana pipiens
25. millipede, millepede, milliped
26. mud turtle
27. night snake, Hypsiglena torquata
28. ringneck snake, ring-necked snake, ring snake
29. slug
30. tailed frog, bell toad, ribbed toad, tailed toad, Ascaphus trui
31. terrapin
32. thunder snake, worm snake, Carphophis amoenus
33. tree frog, tree-frog
34. vine snake
35. water snake
36. whiptail, whiptail lizard\\\hline
b) & 1.	American black bear, black bear, Ursus americanus, Euarctos americanus
2. Angora, Angora rabbit
3. Madagascar cat, ring-tailed lemur, Lemur catta
4. Persian cat
5. baboon
6. badger
7. black-footed ferret, ferret, Mustela nigripes
8. chimpanzee, chimp, Pan troglodytes
9. colobus, colobus monkey
10. giant panda, panda, panda bear, coon bear, Ailuropoda melanoleuca
11. gibbon, Hylobates lar
12. gorilla, Gorilla gorilla
13. guenon, guenon monkey
14. howler monkey, howler
15. indri, indris, Indri indri, Indri brevicaudatus
16. langur
17. lesser panda, red panda, panda, bear cat, cat bear, Ailurus fulgens
18.	macaque
19.	marmoset
20.	meerkat, mierkat
21.	mink
22.	orangutan, orang, orangutang, Pongo pygmaeus
23.	otter
24.	patas, hussar monkey, Erythrocebus patas
25.	polecat, fitch, foulmart, foumart, Mustela putorius
26.	proboscis monkey, Nasalis larvatus
27.	raccoon, racoon
28.	siamang, Hylobates syndactylus, Symphalangus syndactylus
29.	skunk, polecat, wood pussy
30.	sloth bear, Melursus ursinus, Ursus ursinus
31.	spider monkey, Ateles geoffroyi
32.	squirrel monkey, Saimiri sciureus
33.	titi, titi monkey
34.	weasel \\ \hline
c) & 1.	CD player
2. Polaroid camera, Polaroid Land camera
3. camcorder
4. carpenter's kit, tool kit
5. cassette player
6. cellular telephone, cellular phone, cellphone, cell, mobile phone
7. computer keyboard, keypad
8. digital clock
9. flash memory
10. flash, photoflash, flash lamp, flashgun, flashbulb, flash bulb
11. hand calculator, pocket calculator
12. hand-held computer, hand-held microcomputer
13. hard disc, hard disk, fixed disk
14. iPod
15.	laptop, laptop computer
16.	loudspeaker, speaker, speaker unit, loudspeaker system, speaker system
17.	oscilloscope, scope, cathode-ray oscilloscope, CRO
18.	point-and-shoot camera
19.	projector
20.	radio, wireless
21.	reflex camera
22.	remote control, remote
23.	tape player
24.	web site, website, internet site, site \\
 \hline
 \end{tabular}
 \end{minipage}
 \end{table*}
Here we present our clustering results. Due to space constraints, we only present three grouping results based on the CNN feature. Visual effects are shown in \figurename \ref{fig:threeGroups6}, in which each class member is represented by an image. With respect to Group 1, our method obtains similar reptiles. With respect to Group 2, our method acquires most of the monkey classes. Group 3 contains the classes of man-made tools. Our method can group more classes with similar semantic meaning in visual effects. Table 5 provides the class names corresponding to \figurename \ref{fig:threeGroups6}. The CNN feature combined with the proposed similarity metric is distinctive and compact in terms of distinctiveness and generalizability for semantic discrimination.

\begin{figure*}[!t]
\centering
\includegraphics[width=0.96\textwidth]{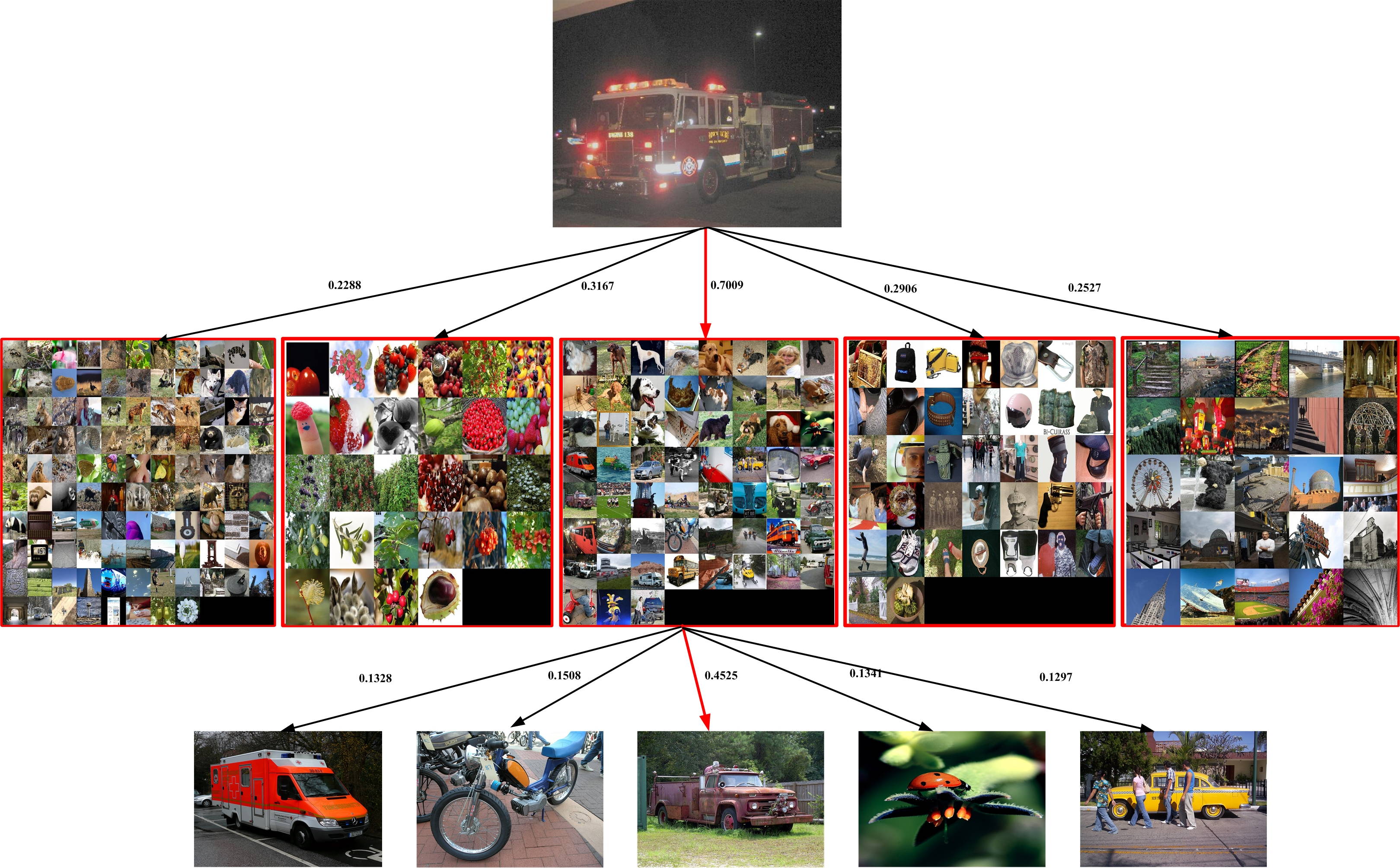}
\caption{An example of the \emph{N-best path} search results on the visual tree $T_{32,2}$ with Inception features on ILSVRC2010.}
\label{fig:fig_5a}
\end{figure*}

\begin{figure*}[!t]
\centering
\includegraphics[width=0.96\textwidth]{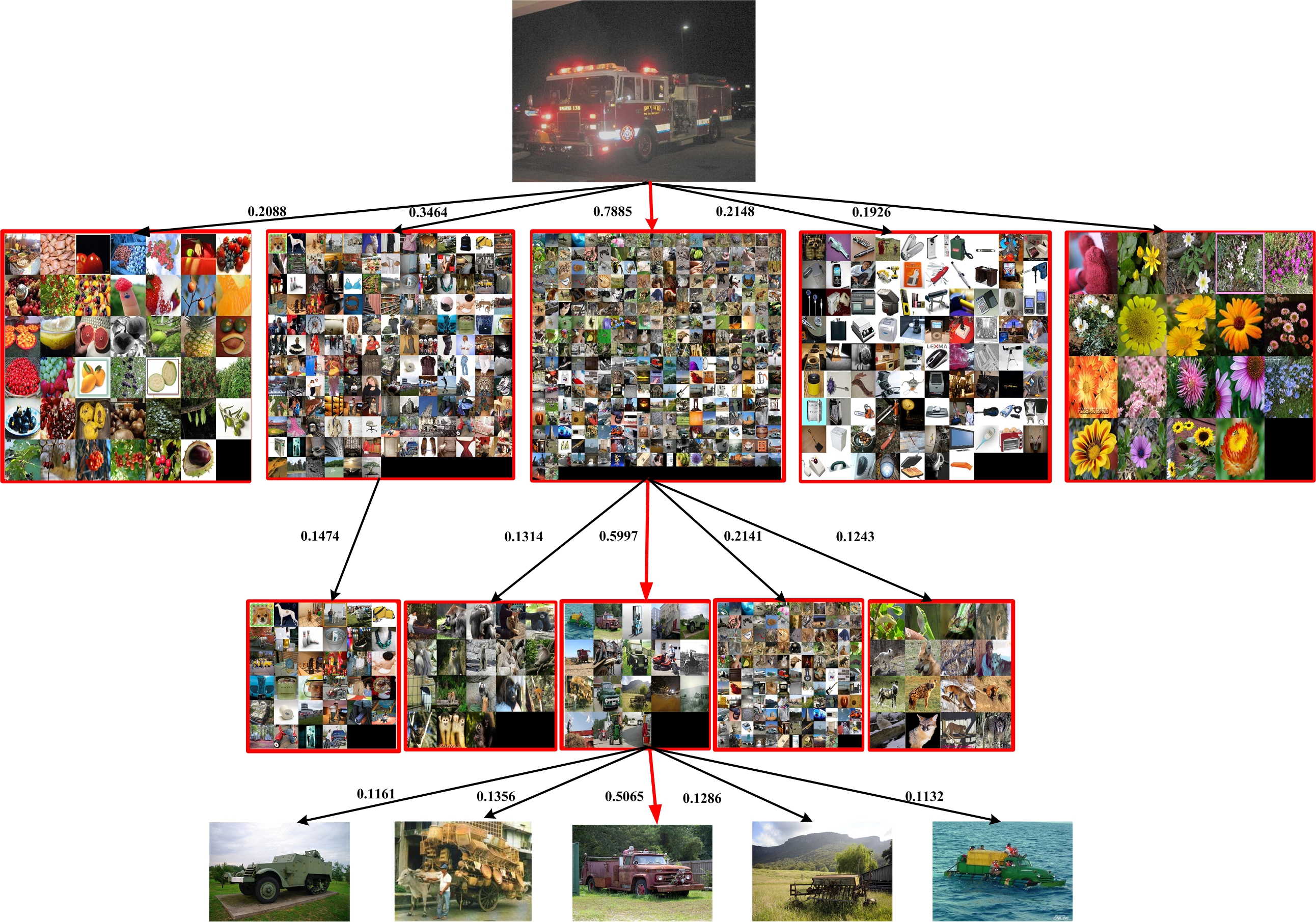}
\caption{An example of the \emph{N-best path} search results on the visual tree $T_{10,3}$ with Inception features on ILSVRC2010.}
\label{fig:fig_5b}
\end{figure*}

We next visualize prediction with the Inception feature (\figurename \ref{fig:fig_5a}, \figurename \ref{fig:fig_5b} ), where  \figurename \ref{fig:fig_5a} shows the path searching process on visual tree $T_{32,2}$, and \figurename \ref{fig:fig_5b} shows the path searching process on visual tree $T_{10,3}$.  For each layer, we show the traversing nodes and its top five edges corresponding to the first $5$ largest probabilities. The optimal path with maximum joint probability is represented with a red thin line with an arrow. By comparing \figurename \ref{fig:fig_5a} and \figurename \ref{fig:fig_5b}, we can see that the groups in the second layer of $T_{32,2}$  are more compact than those in $T_{10,3}$.


\subsection{Deep investigation of visual tree}
 We pay attention to four factors of a visual tree: the structure of the visual tree with different depths and branch number, the number of the visual trees, label prediction scheme and feature representation. In order to investigate the effect of the different depth and branch number, we construct three visual trees: 1) $T_{32,2}$, a visual tree of depth $2$ with no more than $32$ branches per node; 2) $T_{10,3}$, a visual tree of depth $3$ with no more than $10$ branches per node; and 3) $T_{6,4}$, a visual tree of depth $4$ with no more than $6$ branches per node. The reason to choose the three visual trees is that they are the typical tree structures used in previous literatures\cite{5item}\cite{10item}\cite{16item}\cite{30item}. We do not construct very deep visual tree because the error propagation greatly influences the classification performance of the visual tree.

\begin{figure}[!t]
\centering
\includegraphics[width=0.47\textwidth]{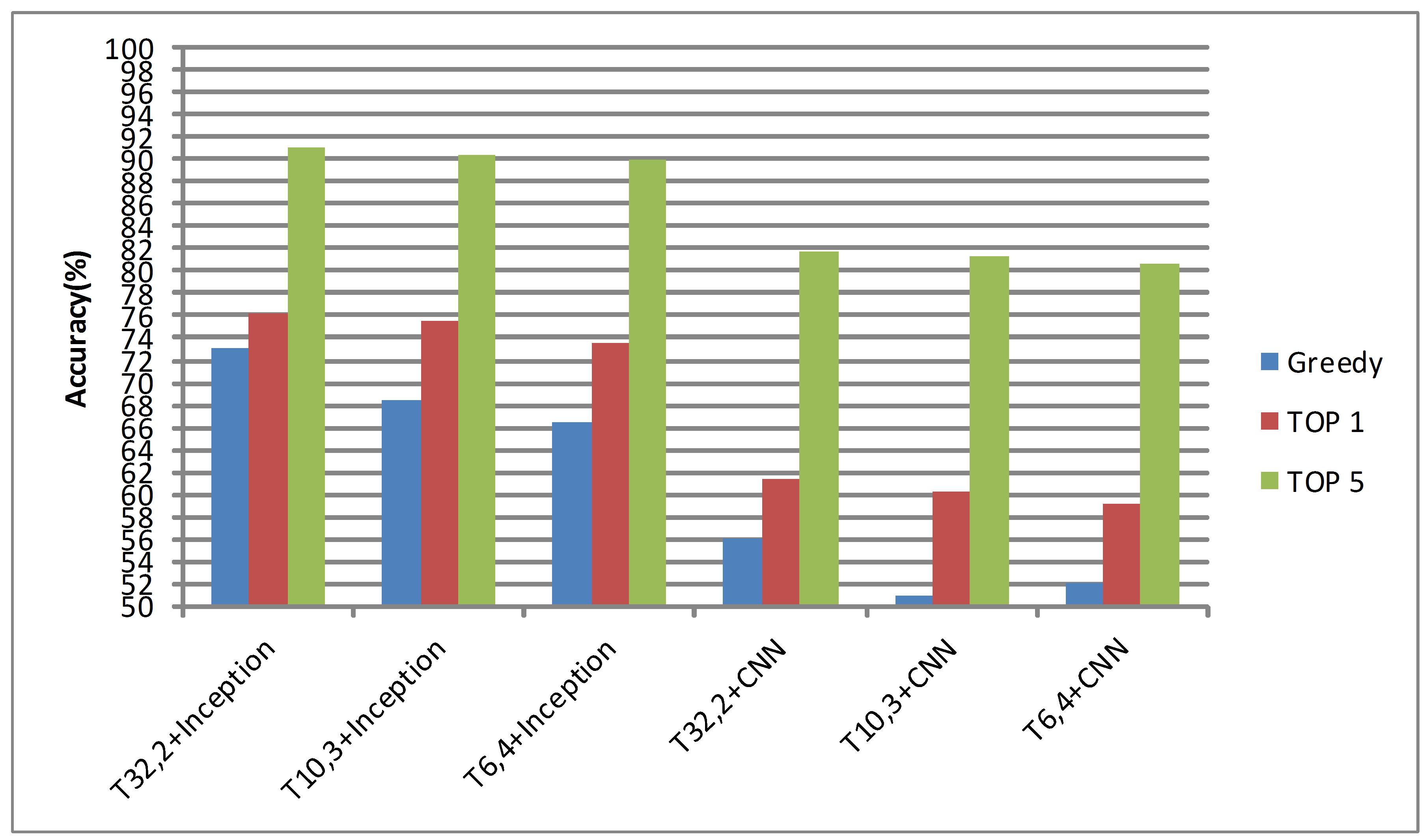}
\caption{Comparison of different visual tree and different label prediction methods on classification accuracy.}
\label{fig:bestpath_vs_greedy}
\end{figure}

 For the label prediction method, we compare the greedy learning method and the N-best path method. In \figurename\ref{fig:bestpath_vs_greedy}, we compare three visual trees with different hierarchical structures and different deep features as well as the prediction method. There are six combinations between the tree structures and the prediction schemes. Each group contains three results, the top-1 classification accuracy with the greedy learning based prediction, and the top-1 and top-5 accuracies with the N-best path prediction. It demonstrates that  $T_{32,2}$  achieves the best of classification accuracy among the visual trees with different depth. It is verified that with the increase of the visual tree depth, the error propagation makes the label prediction poor. For example, under the condition of the Inception feature, the top-1 accuracies are $76.2\%$, $75.6\%$, $73.6\%$ based on the N-best path prediction and $73.0\%$, $68.5\%$, $66.5\%$ based on the greedy learning based prediction corresponding to $T_{32,2}$, $T_{10,3}$, $T_{6,4}$. As for the label prediction method, the N-best path prediction is better than the greedy learning in terms of classification accuracy with the same visual tree. The accuracy differences are $3.2\%$, $7.1\%$, $7.1\%$ between the N-best path prediction and the greedy learning based prediction corresponding to $T_{32,2}$, $T_{10,3}$, $T_{6,4}$. Comparing the two deep features, the Inception feature is more distinctive and achieves better classification performance than the CNN feature, which is the same as the conclusion made in Subsection A.

\begin{figure}[!t]
\centering
\includegraphics[width=0.47\textwidth]{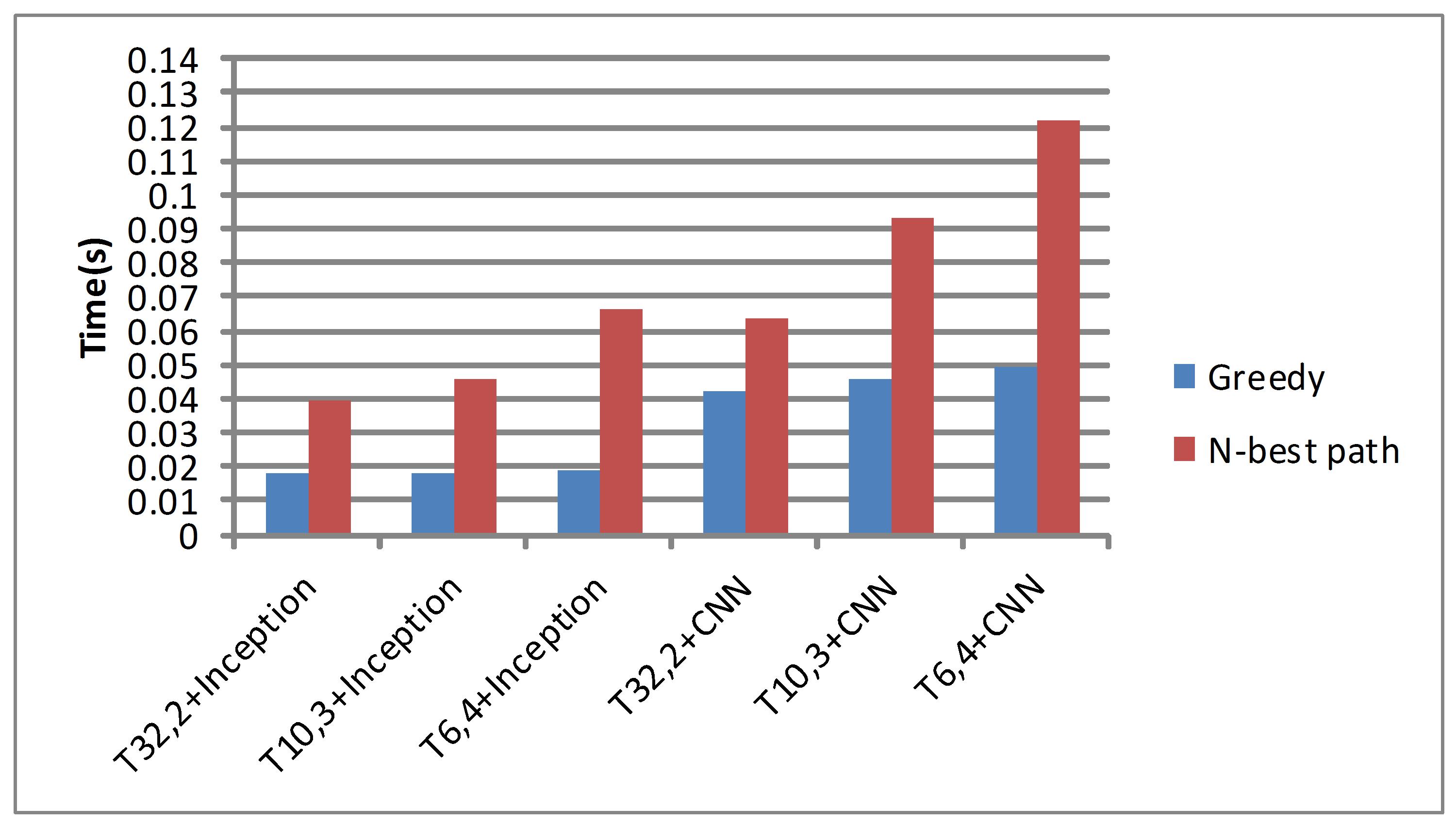}
\caption{Comparison of different visual tree and different label prediction methods on running time.}
\label{fig:timecomparison}
\end{figure}

We further compare the running time of different hierarchical structures. \figurename \ref{fig:timecomparison} shows the comparison results. Greedy learning is a little faster than the N-best path method, and the time differences between them are $(21.5ms, 47.5ms, 71.9ms)$ for the CNN feature, and $(22.0ms, 27.2ms, 48.0ms)$ for the Inception feature corresponding to $T_{32,2}$, $T_{10,3}$, and $T_{6,4}$. However, observing \figurename\ref{fig:bestpath_vs_greedy}, the N-best label prediction is much better than the greedy learning method on classification accuracy.

\begin{figure}[!t]
\centering
\includegraphics[width=0.47\textwidth]{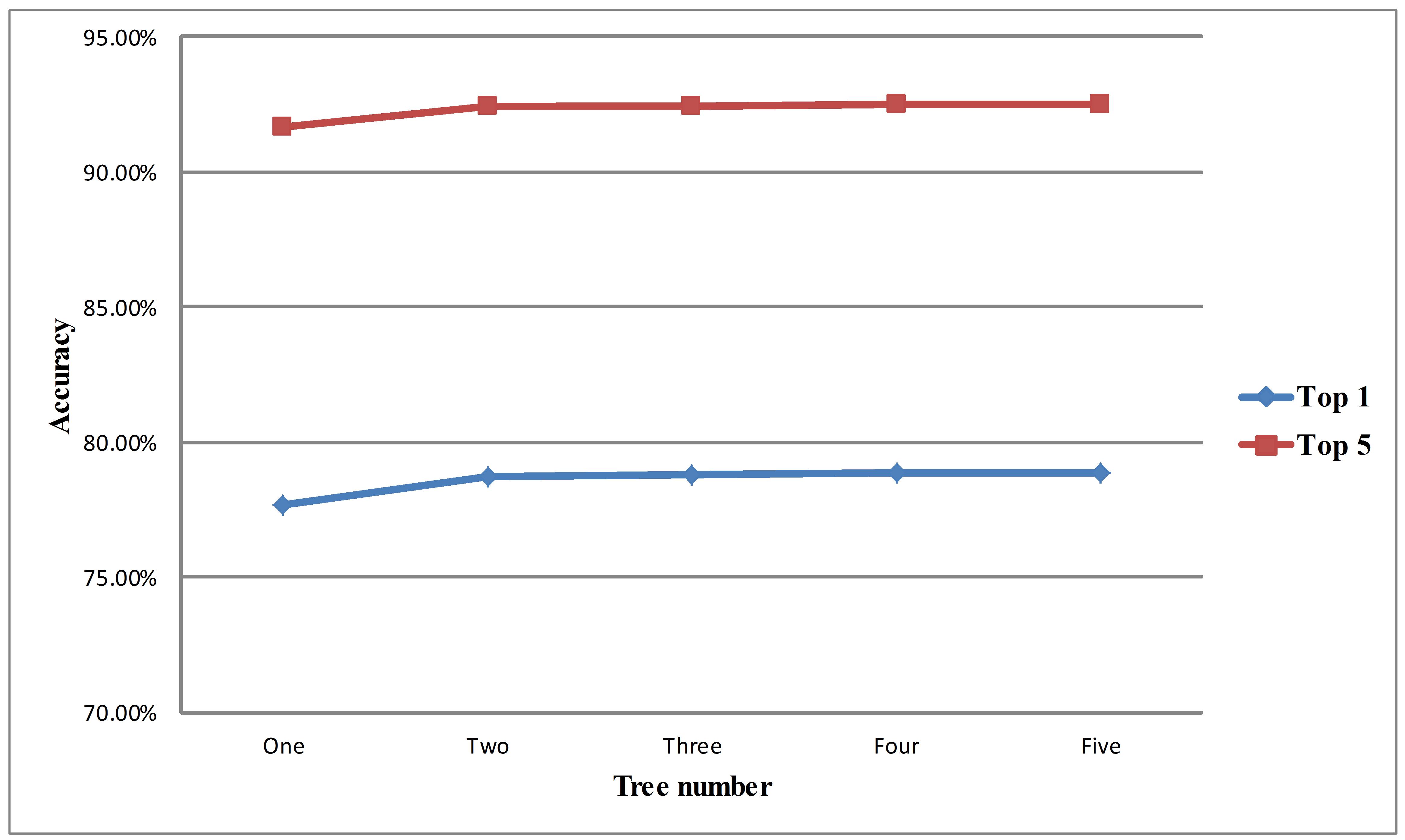}
\caption{The effect of different number of visual tree on classification accuracy.}
\label{fig:treenumbercomparison}
\end{figure}

Furthermore, we have done experiments to investigate the effect of the different number of visual trees in \figurename \ref{fig:treenumbercomparison}. With the increase of the number of the visual tree, the classification accuracy becomes higher, but the increase trend is flat when the number of the visual tree is greater than two. In our experiments, we use five visual trees as the multi-class classification ensemble considering the trade-off  between classification accuracy and computational complexity.

\begin{figure}[!t]
\centering
\includegraphics[width=0.47\textwidth]{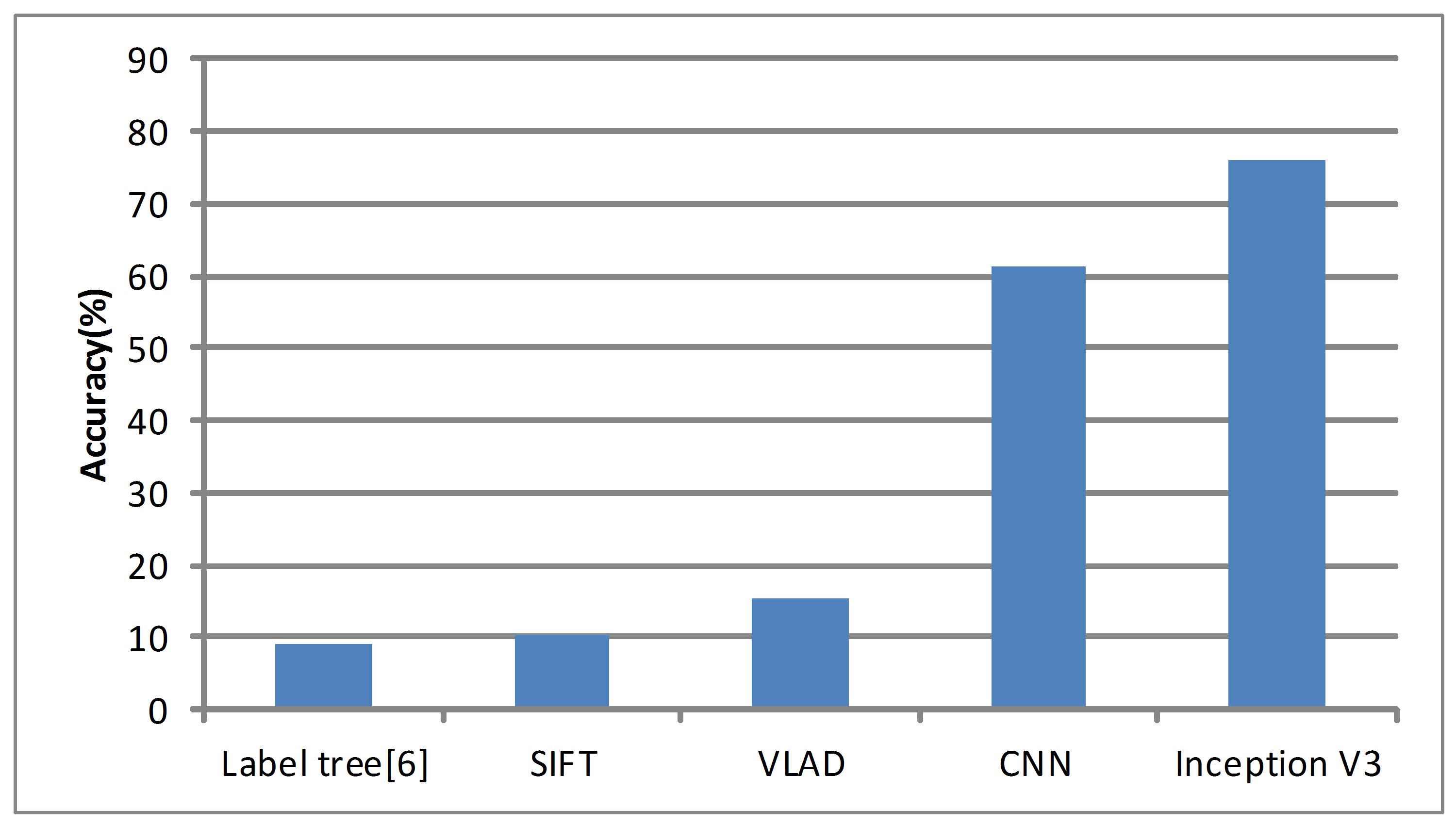}
\caption{Comparison of different features combined with the visual tree $T_{32,2}$ and the label tree together with SIFT.}
\label{fig:comparisonResults8}
\end{figure}

Finally, we investigate the effect of different features combined with the visual tree $T_{32,2}$ on classification accuracy.
 We compare four features: SIFT, VLAD, CNN, and Inception V3. SIFT is downloaded from \cite{1item}, and we use a visual codebook of $1000$ visual terms for image representation. For VLAD, we use $256$ visual words to form a VLAD feature for image representation. Since Label Tree \cite{5item} uses only simple features, we only compare our approach with Label Tree \cite{5item} for fair comparison.
 Results are shown in \figurename \ref{fig:comparisonResults8}. Under the condition of the same feature, our method outperforms Label Tree \cite{5item}, and our approach can achieve even better performance with more distinctive deep features, which is the same as the conclusion in Subsection A.


\subsection{Transferring ability}
\begin{figure*}[!t]
\centering
\includegraphics[width=0.96\textwidth]{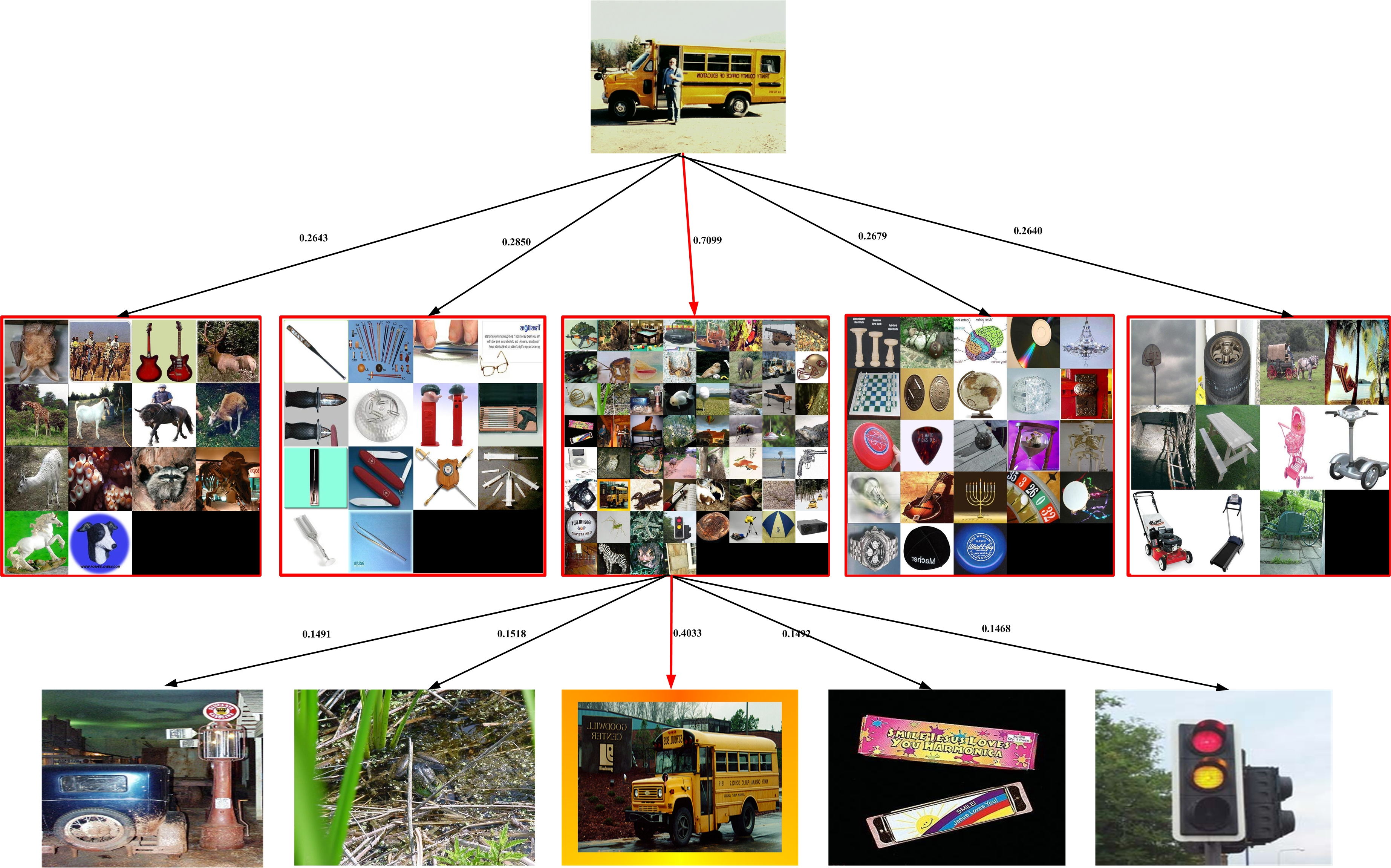}
\caption{An example of our \emph{N-best path} search result on Caltech 256. The weight on an edge is the classification confidence.}
\label{fig:caltech9}
\end{figure*}

We also compare our method with state-of-the-art methods on visual tree $T_{16,2}$ on Caltech 256. To compare our method fairly with other state-of-the-art methods, we use a similar experimental setup. For each category, we randomly sample $N_{train}$ images as the training data and $N_{test}$ images as the test data. Here, $N_{train}=15,30,45$ and $N_{test}=20,30,rest$, where ¡°rest¡± means that the remaining samples except the training samples are used as test data, since similar parameter settings are considered in the most related works. We run our method three times on Caltech 256 with each combination $(N_{train},N_{test} )$. Three groups of data are randomly generated for testing, and the experimental results reported are averages of these three experiments. Overall, the best path search is better than greedy learning in terms of multiple classifications. With an increase in training data, the classification accuracy generally improves.  \figurename \ref{fig:caltech9} shows an example of class prediction inferred by our method on $T_{16,2}$ with CNN features.

Comprehensive comparisons are presented in Table \ref{table 6}. Our approach achieves the best performance on Caltech 256. The reasons for this performance improvement are three-fold: 1) deep features are more distinctive for image representation than traditional feature descriptors such as SIFT; 2) the hierarchical structure is helpful for image classification in addition to the efficiency gain; and 3) our visual tree model  - which combines visual clustering with object classification - is better than the tree model, which only focuses on classification.
\begin{table}[!t]
\renewcommand{\arraystretch}{1.3}
\caption{Comparison with the state-of-the-art methods on Caltech 256 in terms of classification accuracy (\%)}\label{table 6}
\begin{center}
\tabcolsep=5pt
\begin{tabular}{@{}l|cccccc}
\hline
\multirow{2}{*}{Method} & \multicolumn{6}{c}{Feature}\\\cline{2-7}

                              &   \makebox{15}    &   \makebox{30}    &   \makebox{40}    &   \makebox{45}    &   \makebox{50}    &   \makebox{60}    \\ \hline

Griffin\cite{36item}          &   28.3            &	34.1            &     --	        &    --             &	--              &	--              \\ \hline
Gemert \cite{37item}          &   --              &   27.2	        &     --	        &    --             &	--              &	--              \\ \hline
Naveen Kulkarni \cite{38item} &   39.4            &	45.8            &	  --            &	49.3            &	--	            &   51.4            \\ \hline
Yang et al\cite{23item}       &   27.7            &   34.0            &	  --            &	37.5            &	--              &	40.1            \\ \hline
Wang et al\cite{24item}       &   34.4            &   41.2            &     --            &   45.3            &	--              &	47.7            \\ \hline
CRBM\cite{39item}             &   35.1            &   42.1            &	  --            &	45.7            &	--              &	47.9            \\ \hline
N best path\cite{4item}       &   --              &   35.4            &     --            &	 --             &	--              &   --              \\ \hline
Gehler\cite{40item}           &   --              &	45.8            &	  --            &	 --             &	50.8            &   --              \\ \hline
Takumi \cite{41item}          &   40.1            &	48.6            &	  51.6          &	 --             &	53.8            &   --              \\ \hline
CNN ST-BP                          &   64.1            &	68.4            &	  70.1          &	 --             &	--              &   --              \\ 
\hline
Inception3 ST-BP             &   78.7	               &81.3		         &    82.5           &	 --             &	--              &   --              \\ 
\hline
\end{tabular}
\end{center}
\end{table}

We also consider four CNN feature variants: Fisher vector based on CNN features \cite{50item} (Pool$_{5}$+FV), CNN pre-trained on ILSVRC2012 \cite{49item} (ImageNet-CNN), CNN trained on the Places dataset and Caltech 256 \cite{49item} (Hybrid-CNN), and CNN with accurate networks from the Overfeat package \cite{48item}(CNN$_{s}$). We compare the four CNN feature variants with our approach. ImageNet-CNN and hybrid-CNN are similar to our approach and uses a one-vs.-all SVM classifier, while our approach adopts the hierarchical model. The results shown in Table \ref{table 7} suggest that the hierarchical method is superior to the flat classification methods. Furthermore, the deep feature can be improved if it can be extended with more discriminant features, such as Pool$_{5}$+FV. CNN$_{s}$ and Inception V3 which tune the neural network structure improves the classification performance, and Inception V3 achieves the best classification performance compared to the other CNN features on Caltech 256.
\begin{table}[!t]
\renewcommand{\arraystretch}{1.3}
\caption{The comparison of different deep features in terms of classification accuracy (\%) on Caltech256}
\label{table 7}
\begin{center}
\begin{tabular}{l|c}
\hline
    &  Accuracy
 \\ \hline
Pool$_{5}$+FV \cite{50item}  & 79.5 \\ \hline
CNN$_{s}$ \cite{48item}  &     77.6 \\ \hline
ImageNet-CNN \cite{49item}&  67.2 \\ \hline
Hybrid-CNN \cite{49item}  & 65.1 \\ \hline
CNN ST-BP  &  70.1 \\ \hline
Inception3 ST-BP & 82.5
 \\ \hline
\end{tabular}
\end{center}
\end{table}

\section{Conclusions}\label{sec:conclusions}
Here we investigated large-scale object categorization. We proposed a novel multi-class classification framework based on hierarchical category structure learning. The aim of our approach was to improve the efficiency and accuracy of large-scale object categorization with large numbers of multiple classes. The core of our approach was to construct a hierarchical visual tree and to make class predictions based on the visual tree model.  In particular, we constructed the visual hierarchical tree using a fast inter-class similarity computational algorithm and hierarchical spectral clustering. We also proposed an effective path-searching algorithm named \emph{N-best path} for class prediction, which was implemented by a joint probability maximization problem. We evaluated our approach on two large benchmark datasets: ILSVRC2010 and Caltech 256.  The experimental results demonstrated that our method is superior to other state-of-the-art hierarchical learning methods in terms of both the resulting visual tree hierarchy and classification accuracy.

\ifCLASSOPTIONcaptionsoff
  \newpage
\fi
\small{
\section{Acknowledgements}
The authors would like to thank editor and anonymous reviewers who gave valuable suggestions that have helped to improve the quality of the paper. This work was supported by the National Natural Science Foundation of China under Grant 61373077, Grant 61402480, Grant 61502081, in part by the Hong Kong Scholar Program, by Australian Research Council under Grant FT-130101457, DP-140102164 and LE-140100061 and by JSPS KAKENHI under Grant 15K16024.
}




\begin{thebibliography}{1}

\bibitem{1item}	http://www.image-net.org/.
\bibitem{29item} http://caffe.berkeleyvision.org/.
\bibitem{31item} http://www.csie.ntu.edu.tw/~cjlin/liblinear/.
\bibitem{51item} S. Bahrampour, N. M. Nasrabadi, A. Ray, and W. K. Jenkins, ``Multimodal task-driven dictionary learning for image classification,'' \emph{IEEE Trans. Image Processing}, vol. 25, no.1, pp. 24-38, 2016.
\bibitem{14item} E. Bart, I. Porteous, P. Perona and M. Welling, ``Unsupervised learning of visual taxonomies,''
    in \emph{Proc. IEEE CVPR}, June 2008, pp. 1 - 8.
\bibitem{5item} S. Bengio, J. Weston, and D. Grangier, ``Label embedding trees for large multi-class tasks,''
    In \emph{Proc. NIPS, 2010,} pp. 163-171.
\bibitem{48item} K. Chatfield, K. Simonyan, A. Vedaldi, and A. Zisserman, ``Return of the devil in the details: Delving deep into convolutional
    nets,''  In \emph{Proc. BMVC}, 2014.
\bibitem{45item} C. Chiang, C. H. Liu, C. H. Duan, and S. H. Lai, ``Learning component-level sparse representation for image and video categorization,'' \emph{IEEE Trans. Image Process}, vol. 22, no. 12, pp. 4775-4787, 2013.
\bibitem{8item}	G. Csurka, C. R. Dance, L. Fan, J. Willamowski, and C. Bray, ``Visual categorization with bags of keypoints,''
In \emph{Workshop on Statistical Learning in Computer Vision ECCV}, 2004, pp. 1-22.
\bibitem{22item} N. Dalal,and B. Triggs, ``Histograms of oriented gradients for human detection,''
In \emph{Proc. CVPR}, June 2005, pp. 886-893.b
\bibitem{16item} J. Deng, S. Satheesh, A. C. Berg, and F. Li, ``Fast and balanced: Efficient label tree learning for large scale object recognition,''
In \emph{Proc. NIPS}, 2011, pp. 567-575.
\bibitem{17item} P. Dong, K. Mei, N. Zheng, H. Lei, and J. Fan, ``Training inter-related classifiers for automatic image classification and annotation,'' \emph{Pattern Recognition}, vol. 46, no. 5, pp. 1382-1395, 2013.
\bibitem{12item} J. Fan, X. He, N. Zhou, J. Peng, and R. Jain, ``Quantitative characterization of semantic gaps for learning complexity estimation and inference model selection,'' \emph{IEEE Trans. Multimedia}, vol. 14, no. 5, pp. 1414-1428, 2012.
\bibitem{11item}J. Fan, Y. Shen, C. Yang, and N. Zhou, ``Structured max-margin learning for inter-related classifier training and multilabel image annotation,'' \emph{IEEE Trans. Image Process}, vol. 20, no. 3, pp. 837-854, 2011.
\bibitem{30item} J. Fan, J. Zhang, K. Mei, J. Peng, and L. Gao, ``Cost-sensitive learning of hierarchical tree classifiers for large-scale image classification and novel category detection,'' \emph{Pattern Recognition}, vol. 48 no. 5, pp. 1673-1687, 2015.
\bibitem{42item} J. Fan, N. Zhou, J. Peng and Y. Gao, ``Hierarchical learning of tree classifiers for large-scale plant species identification,'' \emph{IEEE Trans. Image Process}, vol. 24, no. 11, pp. 4172-4184, 2015.
\bibitem{50item} B. Gao, X. Wei, J. Wu, and W. Lin, ``Deep Spatial Pyramid: The Devil is Once Again in the Details,'' \emph{CoRR abs}/1504.05277, 2015.
\bibitem{44item} S. Gao, W. Tsang, and Y. Ma, ``Learning category-specific dictionary and shared dictionary for fine-grained image categorization,'' \emph{IEEE Trans. Image Process}, vol.23, no. 2, pp. 623-634, 2014.
\bibitem{7item} T. Gao, and D. Koller, ``Discriminative learning of relaxed hierarchy for large-scale visual recognition,''
in \emph{Proc. IEEE ICCV}, Nov. 2011, pp. 2072-2079.
\bibitem{40item} P. Gehler, and S. Novazin, ``On feature combination for multiclass object classification,''
In \emph{Proc. IEEE ICCV}, Oct. 2009, pp. 221-228.
\bibitem{37item} J. van Gemert, J. Geusebroek, C. Veenman, and A. Smeulders, ``Kernel codebooks for scene categorization,''
In \emph{Proc. ECCV}, 2008, pp. 696-709.
\bibitem{19item} K. Grauman, and T. Darrell, ``The pyramid match kernel: Discriminative classification with sets of image features,''
In \emph{Proc. IEEE ICCV}, Oct. 2005, 1458-1465.
\bibitem{36item} G. Griffin, A. Holub, and P. Perona, ``Caltech-256 object category dataset,'' 2007.
\bibitem{15item} G. Griffin, and P. Perona, ``Learning and using taxonomies for fast visual categorization,''
in \emph{Proc. IEEE CVPR}, June 2008, pp. 1 - 8.
\bibitem{28item} A. Krizhevsky, I. Sutskever, and G.E. Hinton, ``Imagenet classification with deep convolutional neural networks,''
In \emph{Proc. NIPS}, 2012.
\bibitem{41item} T. Kobayashi, ``BOF meet HOG: feature extraction based on histograms of oriented pdf gradients for image classification,''
In \emph{Proc. IEEE CVPR}, June 2013, pp. 747-754.
\bibitem{38item} N. Kulkarni, and B. Li, ``Discriminative affine sparse codes for image classification,''
In \emph{Proc. IEEE CVPR}, June 2011, pp. 1609-1616.
\bibitem{27item} Y. Lecun, B. Boser, J. S. Denker, D. Henderson, R. E. Howard, H. Hubbard, and L. D. Jackel, ``Handwritten digit recognition with a back-propagation network,'' In \emph{Proc. NIPS}, 1997, pp. 396-404.
\bibitem{3item} H. Lei, K. Mei, N. Zheng, P. Dong, N. Zhou, and J. Fan, ``Learning group-based dictionaries for discriminative image representation,''
\emph{Pattern Recognition}, vol. 47, no. 2, pp. 899-913, 2014.
\bibitem{52item} X. Li, X. Zhao, Z. Zhang, F. Wu, Y. Zhuang, J. Wang, and X. Li, ``Joint multilabel classification with vommunity-aware label graph learning, ''\emph{IEEE Trans. Image Processing}, vol. 25, no. 1, pp. 484- 493, 2016.
\bibitem{04item} Y. Li, X. Shi, C. Du, Y. Liu, and Y. Wen£¬
``Manifold regularized multi-view feature selection for social image annotation,'' \emph{Neurocomputing}, vol. 204, pp. 135-141, 2016.
\bibitem{55item} Y. Lin, F. Lv, S. Zhu, M. Yang, T. Cour, K. Yu, L. Cao, and T. Huang, ``Large-scale image classification: Fast feature extraction and SVM training,'' In \emph{Proc. IEEE CVPR}, June 2011, pp. 1689-1696.
\bibitem{10item} B. Liu, F. Sadeghi, M. Tappen, O. Shamir, and C. Liu, ``Probabilistic label trees for efficient large scale image classification,''
in \emph{Proc. IEEE CVPR}, June 2013, pp. 843 - 850.
\bibitem{00item} T. Liu and D. Tao, ``Classification with Noisy Labels by Importance Reweighting'', \emph{IEEE Trans. Pattern Anal. Mach. Intell.}, vol. 38, no. 3, pp. 447-461, March 2016.
\bibitem{21item} D. G. Lowe, ``Distinctive image features from scale-invariant keypoints,''
\emph{International Journal of Computer Vision}, vol.60, no. 2, pp. 91-110, 2004.
\bibitem{06item} Y. Luo, D. Tao, K. Ramamohanarao, C. Xu, and Y. Wen,
''Tensor Canonical Correlation Analysis for Multi-View Dimension Reduction,'' \emph{IEEE Trans. Knowl. Data Eng.}, vol. 27, no. 11, pp. 3111-3124, 2015.
\bibitem{03item} Y. Luo, D. Tao, C. Xu, C. Xu, H. Liu, and Y. Wen,
'' Multiview Vector-Valued Manifold Regularization for Multilabel Image Classification,'' \emph{IEEE Trans. Neural Netw. Learning Syst}, vol. 24, no. 5, pp. 709-722, 2013.
\bibitem{05item} Y. Luo, Y. Wen, and D. Tao, ''On Combining Side Information and Unlabeled Data for Heterogeneous Multi-task Metric Learning,'' In \emph{Proc. IJCAI}, 2016.
\bibitem{54item} Y. Luo, Y. Wen, D. Tao, J. Gui, and C. Xu, ``Large margin multi-modal multi-task feature extraction for image classification,'' \emph{IEEE Trans. Image Processing}, vol. 25, no. 1, pp.414-427, 2016.
\bibitem{18item} M. Marszalek, and C. Schmid, ``Constructing category hierarchies for visual recognition,''
In \emph{Proc. ECCV}, 2008, pp. 479-491.
\bibitem{2item} G. Miller, and C. Fellbaum,``Wordnet: An electronic lexical database,'' 1998.
\bibitem{47item} A. Y. Ng, M. I. Jordan, and Y. Weiss, ``On spectral clustering: Analysis and an algorithm,'' In \emph{Proc. NIPS}, 2002, pp. 849-856.
\bibitem{53item} H. V. Nguyen, H. T. Ho, V. M. Patel, and R. Chellappa, ``DASH-N: joint hierarchical domain adaptation and feature learning,'' \emph{IEEE Trans. Image Processing}, vol.24, no.12, pp. 5479-5491, 2015.
\bibitem{34item} F. Perronnin, Z. Akata, Z. Harchaoui, and C. Schmid, ``Towards good practice in large-scale visual image classification,''
In \emph{Proc. IEEE CVPR}, 2012, pp. 3482 - 3489.
\bibitem{26item} F. Perronnin, J. S¨¢nchez, and T. Mensink, ``Improving the fisher kernel for large-scale image classification,''
In \emph{Proc. ECCV}, 2010, pp. 119-133.
\bibitem{20item} Y. Qu, S. Wu, H. Liu, Y. Xie, and H. Wang, ``Evaluation of local features and classifiers in BOW model for image classification,'' \emph{Multimedia Tools and Applications}, vol. 70, pp. 605-624, 2014.
\bibitem{33item}J. S\'{a}nchez, and F. Perronnin, ``High-dimensional signature compression for large-scale image classification,''
In \emph{Proc. IEEE CVPR}, June 2011, pp. 1665-1672.
\bibitem{58item} F. Shen, C. Shen, Q. Shi, A. Hengel, Z. Tang, and H. Shen, ``Hashing on Nonlinear Manifolds'', \emph{IEEE Trans. Image Processing}, vol. 24, no. 6, pp.1839-1851, 2015.
\bibitem{07item} F. Shen, C. Shen, X. Zhou, Y. Yang and H. Shen, ``Face Image Classification by Pooling Raw Features '', \emph{Pattern Recognition}, vol. 54, pp.94-103, 2016.
\bibitem{43item} L. Shen, G. Sun, Q. Huang, S. Wang, Z. Lin, and E. Wu, ``Multi-level discriminative dictionary learning with application to large scale image classification,'' \emph{IEEE Trans. Image Process}, vol. 24, no. 10, pp. 3109-3123, 2015.
\bibitem{13item} J. Sivic, B. C. Russell, A. Zisserman, W. T. Freeman, and A. A. Efros,
 ``Unsupervised discovery of visual object class hierarchies,'' in \emph{Proc. IEEE CVPR}, June 2008, pp. 1063-6919.
\bibitem{39item} K. Sohn, D. Jung, H. Lee, and A. O. Hero III, ``Efficient learning of sparse, distributed, convolutional feature representations for object recognition,'' In \emph{Proc. IEEE ICCV}, Nov. 2011, pp. 2643-2650.
\bibitem{4item} M. Sun, W. Huang, and S. Savarese, ``Find the Best Path: An Efficient and Accurate Classifier for Image Hierarchies,''
in \emph{Proc. IEEE ICCV}, Dec. 2013, pp. 265-272.
\bibitem{46item} C. Szegedy, V. Vanhoucke, S. Ioffe, J. Shlens, and Z. Wojna, ``Rethinking the Inception Architecture for Computer Vision,'' http://arxiv.org/abs/1512.00567v1.
\bibitem{24item} J. Wang, K. Yu, F. lv, T. Huang, and Y. Gong, ``Locality-constrained linear coding for image classification,''
In \emph{Proc. IEEE CVPR}, June 2010, pp. 3360-3367.
\bibitem{9item} J. Winn, A. Criminisi, and T. Minka, ``Object categorization by learned universal visual dictionary,''
in \emph{Proc. IEEE ICCV}, Oct. 2005, pp. 1800-1807.
\bibitem{01item} C. Xu, D. Tao, and C. Xu, ``Multi-view Intact Space Learning'', \emph{IEEE Trans. Pattern Anal. Mach. Intell.}, vol. 37, no. 12, pp. 2531-2544, December 2015.
\bibitem{02item} C. Xu, D. Tao, and C. Xu,``Large-Margin Multi-view Information Bottleneck'', \emph{IEEE Trans. Pattern Anal. Mach. Intell.}, vol. 36, no. 8, pp. 1559-1572, August 2014.
\bibitem{23item} J. Yang, K. Yu, Y. Gong, and T. Huang, ``Linear spatial pyramid matching using sparse coding for image classification,''
In \emph{Proc. IEEE CVPR}, June 2009, pp. 1794-1801.
\bibitem{56item} Y. Zhang, J. Wu, and J. Cai, `` Compact representation for image classification: To choose or to compress? '' In \emph{Proc. IEEE CVPR}, June 2014, pp. 907-914.
\bibitem{49item} B. Zhou, A. Lapedriza, J. Xiao, A. Torralba, and A. Oliva, ``Learning deep features for scene recognition using places
database,'' In \emph{Proc. NIPS}, 2014, pp. 487-495.
\bibitem{6item} N. Zhou, and J. Fan, ``Jointly learning visually correlated dictionaries for large-scale visual recognition applications,''
\emph{IEEE Trans. Pattern Anal. Mach. Intell.}, vol. 36, no. 4, pp. 715-730, 2014.
\bibitem{25item} X. Zhou, K. Yu, T. zhang, and T. Huang, ``Image classification using super-vector coding of local image descriptors,''
In \emph{Proc. ECCV}, 2010, pp. 141-154.

\end{thebibliography}
%

\vspace*{-2\baselineskip}
\begin{IEEEbiography}[{\includegraphics[width=1in,height=1.25in,clip,keepaspectratio]{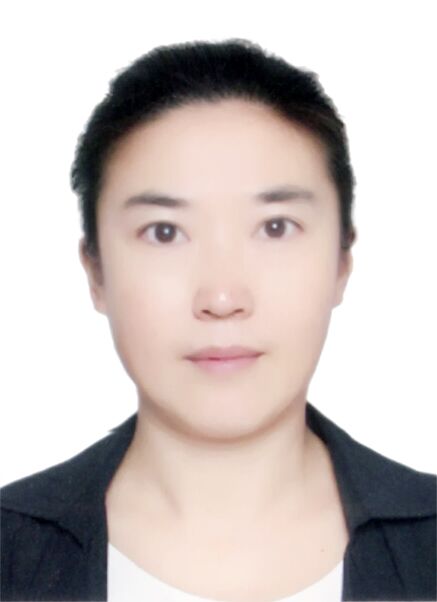}}]{Yanyun Qu}
received the B.S. and the M.S. degrees in Computational Mathematics from Xiamen University and Fudan University, China, in 1995 and 1998, respectively, and received the Ph.D. degrees in Automatic Control from Xi¡¯an Jiaotong University, China, in 2006. She joined the faculty of Department of Computer Science in Xiamen University since 1998. She was appointed as a lecturer from 2000 to 2007 and was appointed as an associate professor since 2007. Her major research interests include pattern recognition and computer vision, with particular interests in large scale image classification and image restoration. She was a technology programme chair of ICIMCS2014. She is a member of IEEE and ACM.
\end{IEEEbiography}

\vspace*{-5\baselineskip}
\begin{IEEEbiography}[{\includegraphics[width=1in,height=1.25in,clip,keepaspectratio]{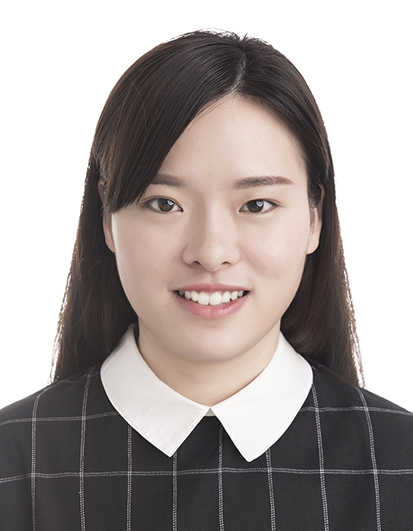}}]{Li Lin}
received the B.Eng. degree from the Department of Computer Science, Xiamen University, in 2014. She is currently working toward the M.S. degree in the Department of Computer Science at Xiamen University. Her research interests include object detection and recognition.
\end{IEEEbiography}

\vspace*{-2\baselineskip}
\begin{IEEEbiography}[{\includegraphics[width=1in,height=1.25in,clip,keepaspectratio]{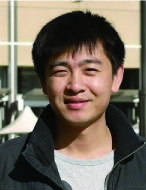}}]{Fumin Shen}
received his B.S. and Ph.D. degree from Shandong University and Nanjing University of Science and Technology, China, in 2007 and 2014, respectively. Currently he is an Associate Professor in school of Computer Science and Engineering, University of Electronic of  Science and Technology  of China, China. His major research interests include computer vision and machine learning, including face recognition, image analysis, hashing methods, and robust statistics with its applications in computer vision. He is a guest editor of Neurocomputing and a special session organizer of MMM'16.
\end{IEEEbiography}
\vspace*{-3\baselineskip}
\begin{IEEEbiography}[{\includegraphics[width=1in,height=1.25in,clip,keepaspectratio]{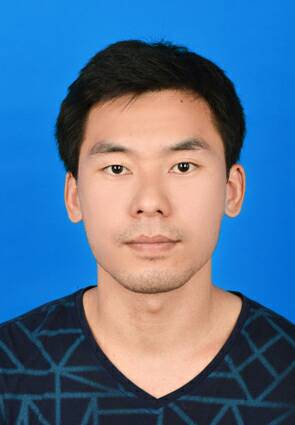}}]{Chang Lu}
received the B.Eng. degree from the Department of Computer Science, Huanggang Normal University in 2013 and the M.S. degree from Xiamen University in 2016. His research interests include object detection and recognition.
\end{IEEEbiography}

\vspace*{-2\baselineskip}
\begin{IEEEbiography}[{\includegraphics[width=1in,height=1.25in,clip,keepaspectratio]{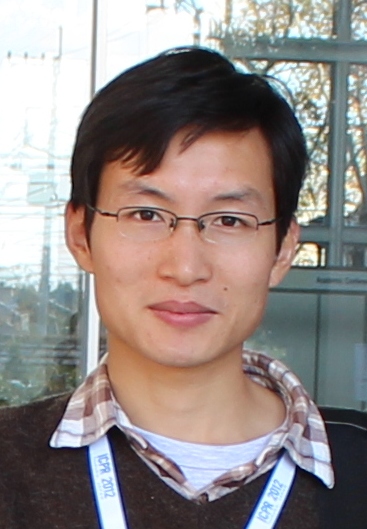}}]{Yang Wu}
received a B.S. degree and a Ph.D degree from Xi'an Jiaotong University in 2004 and 2010, respectively. From Sep. 2007 to Dec. 2008, he was a visiting student in the GRASP lab at University of Pennsylvania. From 2011 to 2014, he was a program specific researcher at the Academic Center for Computing and Media Studies, Kyoto University. Within this period, he was an invited academic visitor at the Big Data Institute of University College London from Jul. 2014 to Aug. 2014. He is currently an assistant professor of the NAIST International Collaborative Laboratory for Robotics Vision, Institute for Research Initiatives, Nara Institute of Science and Technology. His research is in the fields of computer vision, pattern recognition, and image/video search and retrieval, with particular interests in detecting, tracking and recognizing humans and generic objects. He is also interested in pursuing general data analysis models applicable to large data sets.
\end{IEEEbiography}

\vspace*{-2\baselineskip}
\begin{IEEEbiography}[{\includegraphics[width=1in,height=1.25in,clip,keepaspectratio]{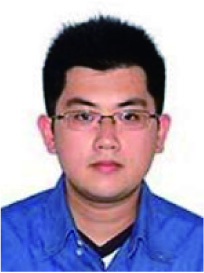}}]{Yuan Xie} (M'12)
received the Ph.D. degree in Pattern Recognition and Intelligent Systems from the Institute of Automation, Chinese Academy of Sciences (CAS), in 2013. He received his master degree in school of Information Science and Technology from Xiamen University, China, in 2010.
He is currently with Visual Computing Laboratory, Department of Computing, The Hong Kong Polytechnic University, Kowloon, Hong Kong, and also with the Research Center of Precision Sensing and Control, Institute of Automation, CAS. He is the author of more than research papers, including more than 20 peer-reviewed articles in international journals such as IEEE Trans. on Image Processing, IEEE Trans. on Neural Network and Learning System, IEEE Trans. on Geoscience and Remote Sensing, IEEE Trans. on Cybernetics, and IEEE Trans. on Circuits and Systems for Video Technology. His research interests include image processing, computer vision, machine learning and pattern recognition.
He received the Hong Kong Scholar Award from the Society of Hong Kong Scholars and the China National Postdoctoral Council in 2014.
\end{IEEEbiography}

\vspace*{-2\baselineskip}
\begin{IEEEbiography}[{\includegraphics[width=1in,height=1.25in,clip,keepaspectratio]{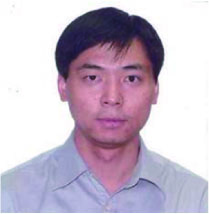}}]{Dacheng Tao} (F'15)
is Professor of Computer Science and Director of the Centre for Artificial Intelligence, and the Faculty of Engineering and Information Technology in the University of Technology Sydney. He mainly applies statistics and mathematics to Artificial Intelligence and Data Science. His research interests spread across computer vision, data science, image processing, machine learning, and video surveillance. His research results have expounded in one monograph and 200+ publications at prestigious journals and prominent conferences, such as IEEE T-PAMI, T-NNLS, T-IP, JMLR, IJCV, NIPS, ICML, CVPR, ICCV, ECCV, AISTATS, ICDM; and ACM SIGKDD, with several best paper awards, such as the best theory/algorithm paper runner up award in IEEE ICDM¡¯07, the best student paper award in IEEE ICDM¡¯13, and the 2014 ICDM 10-year highest-impact paper award. He received the 2015 Australian Scopus-Eureka Prize, the 2015 ACS Gold Disruptor Award and the 2015 UTS Vice-Chancellor¡¯s Medal for Exceptional Research. He is a Fellow of the IEEE, OSA, IAPR and SPIE.
\end{IEEEbiography}

%

%
%
%

\end{document}